\documentclass[journal]{IEEEtran}

\usepackage{cite}
\usepackage{amsmath,amssymb,amsfonts}
\usepackage{graphicx}
\usepackage{multirow}
\usepackage{textcomp}
\usepackage{xcolor}
\usepackage{array}
\usepackage{tikz}
\usepackage{pgfplots}
\pgfplotsset{compat=newest}
\usepackage[utf8]{inputenc}
\usepackage{algorithm}
\usepackage{algpseudocode}
\usepackage{booktabs}
\usepackage{float}

\usepackage{hyperref}

\usepackage{adjustbox}

\def\BibTeX{{\rm B\kern-.05em{\sc i\kern-.025em b}\kern-.08em
    T\kern-.1667em\lower.7ex\hbox{E}\kern-.125emX}}

\newcommand{\newPart}{\textcolor{black}}
\newcommand{\RNum}[1]{\uppercase\expandafter{\romannumeral #1\relax}}

\begin{document}

\title{Using Machine Learning for Anomaly Detection on a System-on-Chip under Gamma Radiation}

\author{Eduardo~Weber~Wächter,
        Server Kasap,~\IEEEmembership{Member,~IEEE}
        \c{S}efki Kolozali,
        Xiaojun Zhai,~\IEEEmembership{Member,~IEEE}
        Shoaib Ehsan,~\IEEEmembership{Senior Member,~IEEE}
        and Klaus McDonald-Maier,~\IEEEmembership{Senior Member,~IEEE}
\thanks{E. Wachter is with the School of Computer Science and Electronic Engineering, University of Essex, Colchester, U.K. and also  with  the School of Engineering, University of Warwick, Coventry, UK}%
\thanks{S. Kasap is with the School of Computing, Electronics and Mathematics, Coventry University, Coventry, UK}%
\thanks{\c{S}. Kolozali, X. Zhai, S. Ehsan and K. McDonald-Maier are with the School of Computer Science and Electronic Engineering, University of Essex, Colchester, U.K.}
}


\maketitle

\begin{abstract}

The emergence of new nanoscale technologies has imposed significant challenges to designing reliable electronic systems in radiation environments. A few  types  of  radiation like Total Ionizing Dose (TID) effects often cause permanent damages on such nanoscale electronic devices, and current state-of-the-art technologies to tackle TID make use of expensive radiation-hardened devices. This paper focuses on a novel and different approach: using machine learning algorithms on consumer electronic level Field Programmable Gate Arrays (FPGAs) to tackle TID effects and monitor them to replace before they stop working. This condition has a research challenge to anticipate when the board results in a total failure due to TID effects. We observed internal measurements of the FPGA boards under gamma radiation and used three different anomaly detection machine learning (ML) algorithms to detect anomalies in the sensor measurements in a gamma-radiated environment. The statistical results show a highly significant relationship between the $\gamma$ radiation exposure levels and the board measurements. Moreover, our anomaly detection results have shown that a One-Class Support Vector Machine with Radial Basis Function Kernel has an average Recall score of 0.95. Also, all anomalies can be detected before the boards stop working.

\end{abstract}

\begin{IEEEkeywords}
Machine Learning, Anomaly Detection, Gamma radiation, Field Programmable Gate Arrays, TID
\end{IEEEkeywords}

\IEEEpeerreviewmaketitle

\section{Introduction}

One of the biggest challenges in the European Union is the cleaning of nuclear waste~\cite{NDA}. This task involves handling and moving extreme toxic material contaminated with different types of ionizing radiation. Unfortunately, higher doses of radiation harm the human body; therefore, the cleaning process should be taken with much precaution. Luckily, most of this work can be done by robots, but sadly the electronic devices controlling them are also susceptible to radiation. Remarkably, many artificial intelligence algorithms used in robots heavily rely on high-speed processors or graphics processing units (GPUs) which leads to significant growth in reliability issues due to the nature of the nanoscale technologies used in those chips, 

One can categorize the radiation consequences on contaminated sites into two broad types: permanent and transient effects. One example of the permanent effects is the Total Ionizing Dose (TID), while Single Event Upset (SEU) is an example of transient effects. TID effect is a phenomenon that causes \textbf{permanent} damage, which inevitably, at some point, is going to result in total failure of the electronics. TID effects can only be minimized to extend the system life. The literature has come up with special hardened devices specifically targeting this effect \cite{Goncalves2013,Ramaswamy09}. Typically, these devices are much more expensive than unhardened ones.
On the other hand, SEUs are \textbf{transient} effects that cause bit-flips on memory elements. For example, one of the most employed solutions for radiation environments, especially for space applications, is Field Programmable Gate Arrays (FPGAs) due to their reconfiguration capability and performance aspects. FPGAs are predominantly used for these applications \cite{lindoso17,Quinn15,Trimberger15} as they allow reconfiguration and thus hardware adaptability to deal with most transient effects.

The underlying idea of this paper is that instead of using an approach with radiation-hardened devices, we adopt consumer electronic level COTS (commercial off-the-shelf) devices and then monitor, manage and replace them with healthy ones before they stop working. This condition brings us a research challenge to anticipate when the board results in a total failure due to radiation, specifically TID effects.

In this paper, we first employ a statistical analysis of the collected data to prove that the measured values of the board under radiation show statistical significance compared to a scenario without radiation. Later, the paper tries to answer whether it is possible to predict when the board will stop working due to radiation with reasonable accuracy. For this purpose, we tested a low-cost COTS unhardened consumer electronic level 28nm FPGA used in many fault-tolerant techniques for transient radiation effects. We employ this board under gamma ($\gamma$) radiation and log its behaviour.

The monitoring of these boards needs to be executed with an enhanced approach. The paper shows that using simple techniques to observe voltage and temperatures may not indicate that the board will stop working. For example, one widely used data analysis tool is R control charts. This type of chart, popularly known as a control chart, monitors the mean and range of normally distributed variables simultaneously when samples are collected at regular intervals. Such a technique uses upper and lower control limits (UCL and LCL) to monitor the behaviour of the variables. Still, it may not be sufficient to indicate that a given board is behaving abnormally. A few data points outside the operational limits caused by radiation do not suggest that the board will stop operating. This paper shows that this might take a few minutes or more than one hour. For this reason, we employ state-of-the-art machine learning (ML) techniques to understand and try to predict when the board is behaving abnormally. \newPart{Authors in \cite{LEUKEL202187} reviews works that used ML for failure prediction in industrial mechanical systems for the last decade and identify opportunities for future research, although there is no focus on electronics.}

The novelty of this works is three-folded: First, this is the first study to monitor and measure voltages and temperatures on a consumer electronic level SRAM-based FPGA SoC (System on Chip) under $\gamma$ radiation. Second, this is the first study performing a quantitative/statistical analysis of the effects of gamma radiation on voltages and temperatures of an FPGA and then compare to an environment without radiation. Finally, this is the first work using machine learning algorithms trying to predict when the board will be rendered in-operational by \textbf{gamma radiation} through the observation of temperature and voltage values only.

The paper is organized as follows. Section~\ref{experimental design} details the hardware setup to monitor and test the boards under experiment, later presenting an example where the proposed technique might be used. Section~\ref{related_works} reviews the state-of-the-art of related techniques. Section~\ref{statistical_effects} explains the statistical analysis while Section~\ref{ML_techniques} details the employed machine learning techniques and how to measure its accuracy. Section~\ref{rad_experiments} details the experiments carried out under $\gamma$ radiation and how the data was organized to feed the machine learning algorithms. Finally, Section~\ref{evaluations} evaluates, compares and discuss the results, and Section~\ref{conclusion} draws conclusions and future plans.

\section{Related Work}
\label{related_works}

As the sophistication of embedded systems grows, their vulnerability to errors is adversely affected due to an increase in critical points of failure. Adopting fault mitigation or fault tolerance techniques is vital if FPGAs are used in radiation environments. Fault tolerance techniques that enhance embedded processor reliability can be categorized as hardware-, software- and hybrid-based techniques~\cite{kasap20}.

The hardware-based techniques, which mainly rely on spatial redundancy, provide two or more instances of a hardware component, such as processors, memories, buses or power supplies, for protection against soft errors. This class of techniques include Triple Modular Redundancy (TMR)~\cite{Xilinx06}, Duplication with Comparison (DWC)~\cite{pradhan97} and hardware monitors~\cite{parra14} which incorporate watchdog or checker modules to monitor the system and detect errors by verifying the control-flow related memory accesses of the target processor. These techniques can protect the system from errors in the computation outputs, i.e. SDCs, as exemplified in~\cite{quinn17}.

Software-implemented hardware fault tolerance (SIHFT) approaches handle hardware malfunctions by merely shielding the software without any hardware alteration. These techniques rely on adding redundant software code for comparison to detect errors. However, they exhibit a high-performance overhead, which may not be viable for some real-time systems. These kinds of techniques, such as ABFT~\cite{huang84}, HETA~\cite{azambuja13} and S-SETA~\cite{chielle15}, detect control-flow faults leading to FIs, which manifest themselves as hangs or crashes, and then place the system into a fail-safe state. \newPart{Note that both hardware- and software-based techniques are not capable of correcting 100\% of the errors}, but rather detecting them to avoid a failure that would have adverse effects on the entire mission. Furthermore, they protect the system either from SDCs or FIs, but not both.

The hybrid techniques are the ones that use a SIHFT method combined with a hardware intellectual property (IP), which performs consistency checks in the processor, making them effective against both SDCs and FIs. For instance, the lockstep technique is a hybrid fault-tolerance technique based on software and hardware redundancy. It employs the concepts of checkpointing and recovery mechanisms (e.g. roll-back recovery, roll-forward recovery) at the software level, and processor replication and checker circuits at the hardware level, as explained in the following sections. Therefore, it is capable of both error detection and correction. The lockstep technique’s most significant merit is its ability to detect and correct both SDCs and FIs, unlike many other fault tolerance techniques. Several researchers have developed and implemented their lockstep technique version, such as those in~\cite{Xilinx07,abate09,violante11,pham13,oliveira18}, to make a range of processors resistant to radiation-induced soft errors, extensively analyzed and compared in~\cite{wachter19}.

The Authors in ~\cite{Rezzak18} evaluated a ($^{60}C$) gamma-ray radiation testing of a space application FPGA, namely the RT4G150 from Microsemi. Microsemi is a qualified manufacturers list (QML)-certified manufacturer of high-reliability FPGAs for space applications, while RTG4 is the 4$^{th}$ generation family of radiation-tolerant flash-based FPGAs. The work assesses the degradation of the flash cell through its threshold-voltage ($V_T$) shift.
For space applications, dynamic burn-in (DBI) testing is used to evaluate the long term reliability of the device. Among all product screening tests employed by many business categories, including automotive, aerospace, and defence, the burn-in (BI) test is one of the most effective tests for early failure detection. The work indicates that RTG4 shows a shift of the programmed Pflash cell $V_T$ post-DBI is observed. The programmed Pflash VTshift is due to voltage degradation, resulting from approximately 1.75\% degradation of the DAC’s output. \newPart{It is essential to emphasize that this work does not monitor the temperature of the FPGA, this being out of the scope of the project.}

In ~\cite{villa2017}  Authors evaluated a COTS FPGA, namely Microsemi ProASIC3E A3PE1500. Despite their low reliability, the authors state that this FPGA has been considered a promising alternative to replace radiation-hardened ones. The paper analyses the Single-Event Upset (SEU) sensitivity of the FPGA for a combined set of Electromagnetic Interference (EMI) and TID tests. This component was under the pre-qualification process for use in some satellites of the Brazilian Space Program. The TID test was performed by exposing the FPGA to a 10-keV effective energy X-ray beam. The device was roughly exposed to the TID expected to be cumulated on satellite electronics after operation for a period of 4 or 5 years in a given orbit, as specified by the Brazilian National Institute for Space Research. The conclusion is that FFs present a  lower  SEU-immunity degradation when exposed to conducted-EMI (5.6\%) than  SRAM  cells  (resp.  6.3\%); the latter memory elements are intrinsically more robust to EMI since they present a much lower cross-section than FFs.  \newPart{Note that this work does not monitor the temperature of the FPGA as well.}

Tarrilho et al.~\cite{Tarrillo2011} analyze the behaviour of flash-based FPGA from Actel under TID. In this work, a design with an embedded system composed of a MIPS microprocessor hardened with fault-tolerant techniques is employed on a COTS flash-based FPGA ProASIC3E family from Actel.
The TID experiment with no reconfiguration monitored the power supply current during radiation and the FPGA temperature. They reported that the Icc started to change after 45 krad(Si) when some modules stopped working. The current increases promptly and reaches 1.5 times the original current just before 65 krad(Si).  The temperature and current drop abruptly when most modules fail around 65 krad(Si). The paper primarily showed the failure dose for some internal modules and did not present when the current starts behaving abnormally.

In ~\cite{Lentaris2019} evaluate TID effects on an SRAM-based COTS FPGA. They combine hardware and software techniques to perform on-chip irradiation via a $^{90}Sr$/$^{90}Y$ electron source and assess the degradation of the system. The experiment consists of two executions of Zynq XC7Z020T chips hosted by two distinct Zedboards. The analysis focuses on ring oscillators (ROs) implemented on the programmable logic of the FPGA for estimating/predicting the performance degradation due to the TID effects. The authors show specifically the RO frequency according to temperature, current and accumulated dose. The authors indicate that specifically for COTS 28-nm Zynq-7000 chips, the results show increased TID tolerance beyond the Mrad level. The tests were conducted with the FPGAs surviving up to 2 Mrad(Si). In summary, this work shows that COTS FPGAs can survive significant amounts of ionizing radiation, although not displaying the limit, since it was not the objective of their work.

These papers~\cite{Tarrillo2011,villa2017,Rezzak18} summarize most related work for TID effects in FPGAs. They either focus on flash-based devices\cite{Tarrillo2011,Rezzak18} or point out \cite{villa2017} that they could be a promising alternative for radiation-hardened ones. Also, only one paper monitors current and temperature, although it is not the main objective of that paper to correlate or predict the FPGA stop working.



\section{Experimental Design }
\label{experimental design}
\begin{figure}[h!]
\centering{\includegraphics[width=1\columnwidth]{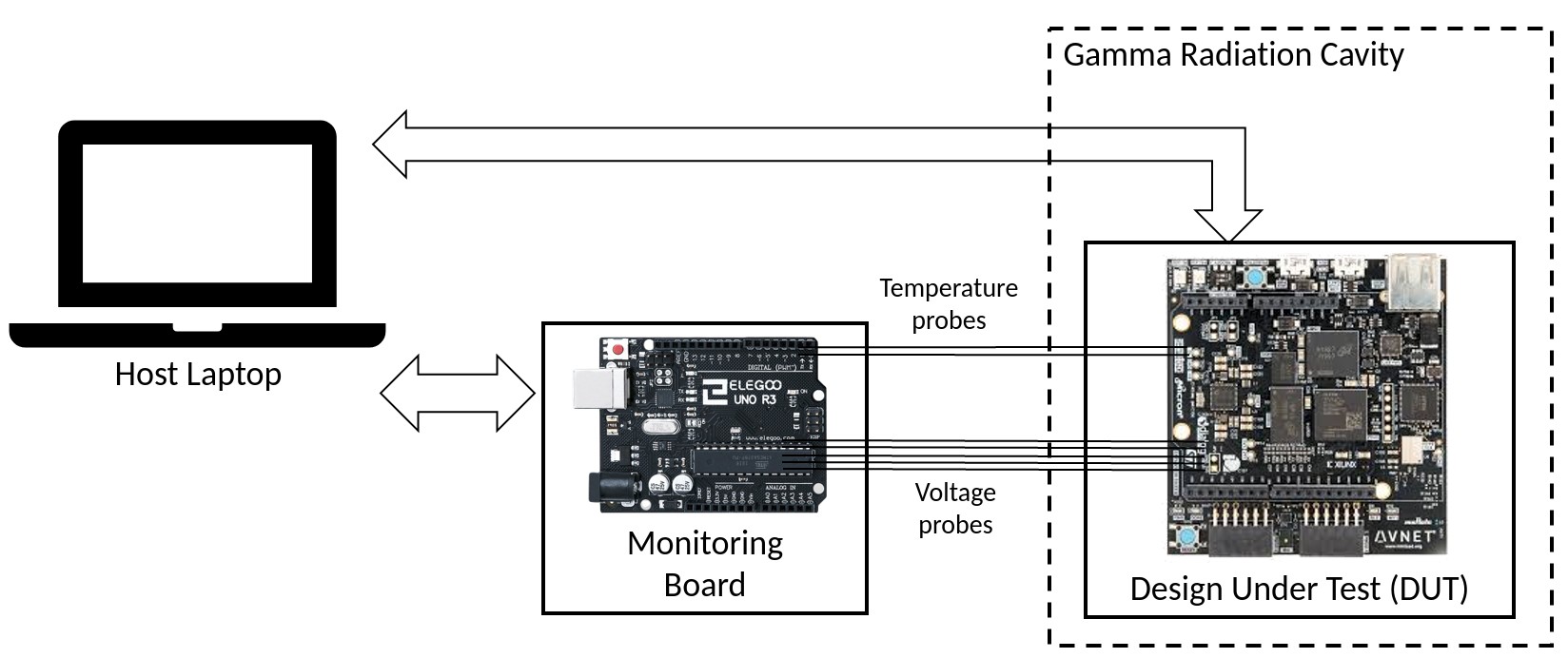}}
\caption{Block diagram of the experimental setup. Only the DUT is under radiation.}
\label{setup}
\end{figure}

This section explains the setup, shown in Fig.~\ref{setup}, built to measure and log all the tested experiments, later detailed in Section~\ref{rad_experiments}. The experimental setup is composed of (i) a laptop to control, collect and log all data; (ii) a \textit{Design Under Test (DUT) board} and (iii) a special \textit{monitoring board} to collect the data from it. The only electronic device exposed to radiation is the DUT, that being the reason for the monitoring board, which is outside the radiation cavity and connected to the DUT with voltage and temperature probes.

\begin{figure}[h!]
\centering{\includegraphics[width=0.8\columnwidth]{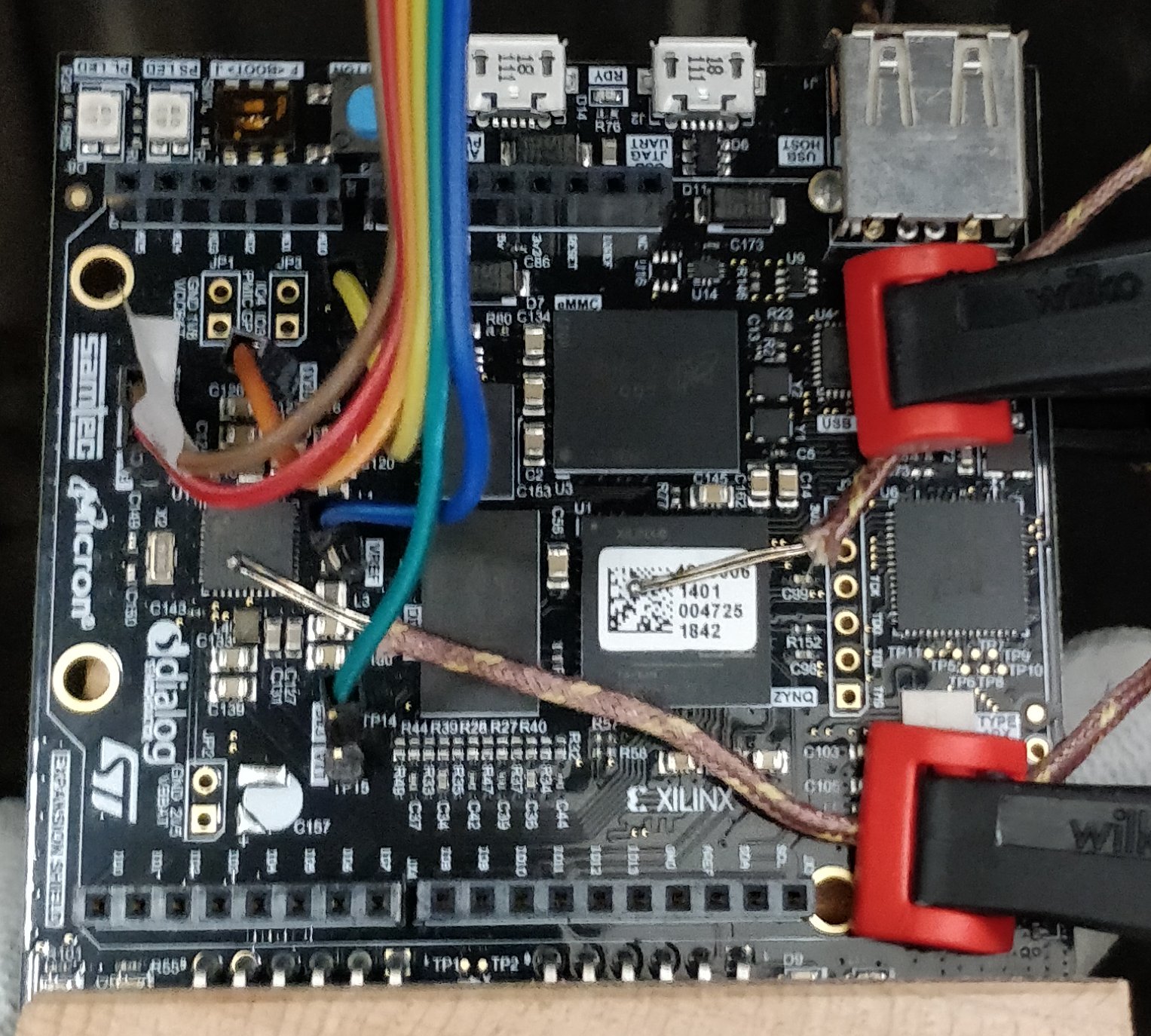}}
\caption{Employed a Minized board with connectors attached to monitor voltage and two thermocouple pairs to measure temperatures.}
\label{minized}
\end{figure}

The DUT is a MiniZed board, depicted in Fig.~\ref{minized}, a widely available and, most used, equipped with a low-cost ($\approx$£100) Xilinx 28nm Zynq FPGA. The monitoring board is an Arduino board coupled with temperature and voltage sensors, which during the experiments are located outside the radiation chamber; therefore, it can be reused.

The MiniZed board \cite{minized} (Fig.~\ref{minized}) is a development board containing a Xilinx Zynq single-core SoC XC7Z007S-1CLG225C \cite{zynq} with 512MB DDR3L micron storage, a 128MB QSPI flash and 8GB eMMC. 
The FPGA contains a programmable logic and a Programmable Logic (PL)(FPGA) and Processing Subsystem(PS) (ARM Cortex-A9). The FPGA chip is fed by a power management integrated circuit (PMIC) Dialog DA9062 \cite{pmic} which controls five different voltages\cite{minized_power}: 0.675, 1.0, 1.35, 1.8 and 3.3 volts, detailed in Table~\ref{table_temp_volt}. 

For example, $V_{DDR3}$ supplies the voltage to the DDR3 memory of the board.
The monitoring board is organized to monitor these voltages. We also use a thermocouple Wire to monitor the temperature on the surface of both FPGA and PMIC. This setup is arranged so that only the DUT and wires connecting to the monitoring board are inside the radiation chamber; therefore, the radiation does not interfere with the monitoring electronics.

\begin{table}[h]
\renewcommand{\arraystretch}{1.3}
\caption{{Temperature and voltage monitored values}}
\label{table_temp_volt}
\centering
\begin{tabular}{|l|l|c|}
 \hline

 \textbf{Symbol} & \textbf{Expected value} & \textbf{Description} \\ [0.5ex]
 \hline

$T_{FPGA}$ & - & FPGA temperature \\\hline
$T_{PMIC}$ & - & PMIC temperature \\\hline
$V_{core}$ & 1 V & Core supply \\\hline 
$V_{aux}$ & 1.8 V & Auxiliary supply \\\hline
$V_{ddr3}$ & 1.35 V & DDR3 \\\hline
$V_{tt}$  & 0.675 V & Vtt \\\hline
$V_{cco}$ & 3.3 V & Vcco / board voltage \\
 \hline
\end{tabular}
\end{table}
\begin{figure}[h!]
\centering{\includegraphics[clip,trim={20cm 0 35cm 0},width=1\columnwidth]{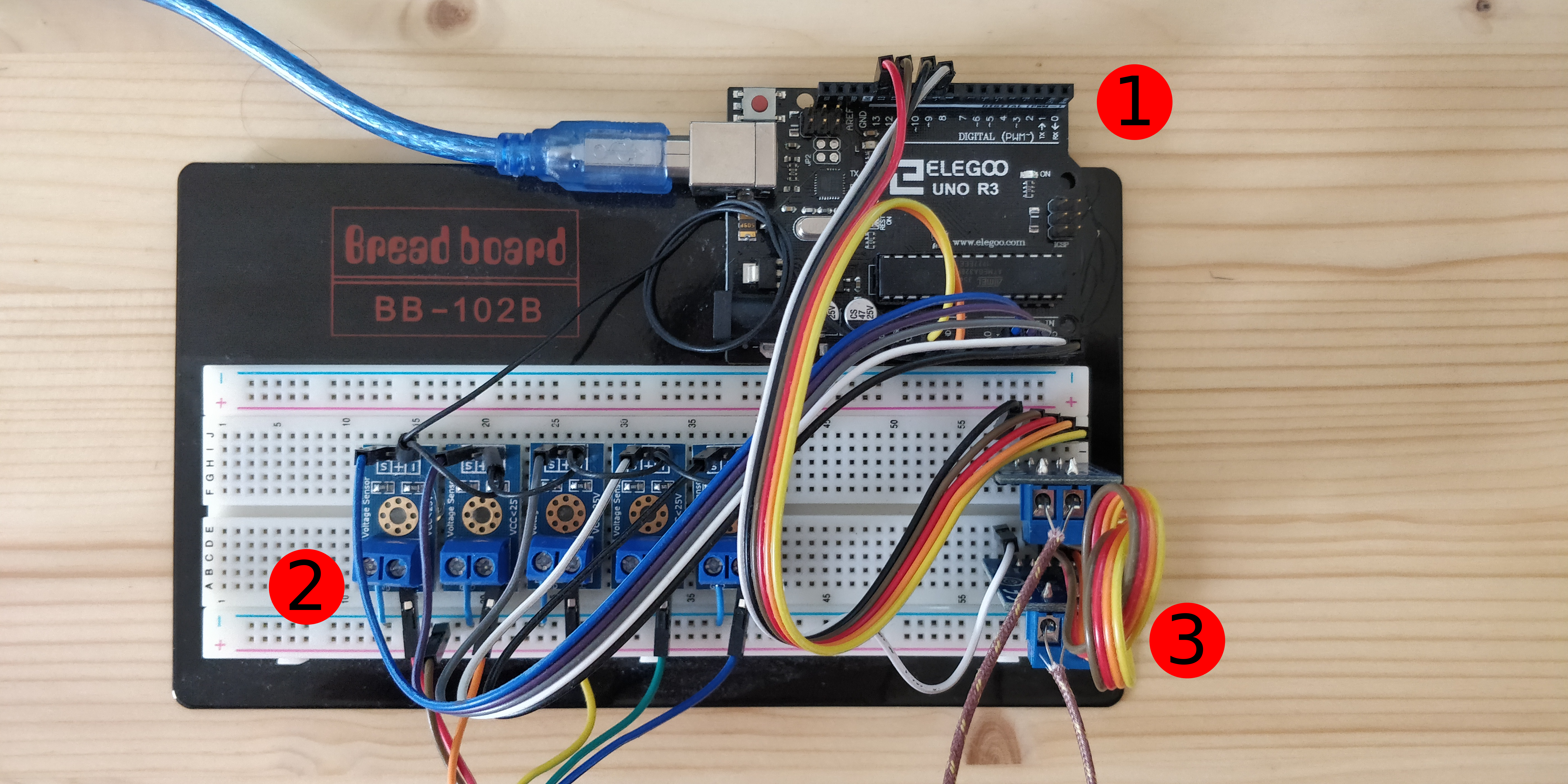}}
\caption{The monitoring board composed of an Elegoo R3 board (1) coupled with voltage (2) and temperature (3) sensors.}
\label{arduino}
\end{figure}

The monitoring board is an Elegoo R3 board \cite{ELEGOO}, which is completely compatible with the official Arduino R3 version. The board is composed of an Atmel ATMEGA328P chip \cite{atmegaDATASHEET} and coupled with five DC0-25V voltage sensors and two Digilent Pmod TC1 K-Type thermocouple modules \cite{PmodTC1}.
The thermocouple sensor module works with a temperature range of -73°C to 482°C with a 0.25°C resolution. The voltage sensor works with a range of DC 0 to 25 V with a resolution of 0.01 V.
The board collects one reading each second, operating in a \textit{1 Hz frequency}. The data is sent to a laptop and stored for later computation.
In parallel, the MiniZed board runs a compute an intense set of operations whose outputs are sent to the host laptop through a serial interface for a sanity check aiming to determine if the board is still operational. When the board stops sending data, it is considered to be dead. Later, all the boards used on experiments were tested individually outside the radiation chamber to ensure they were inoperational. We tried to download the provided sanity check, and none of the boards responded to the cable connection attempts, confirming that the boards were dead.


\subsection{Motivational example}
\label{motivational}

\begin{figure}[htb!]
\centering{\includegraphics[width=1\columnwidth]{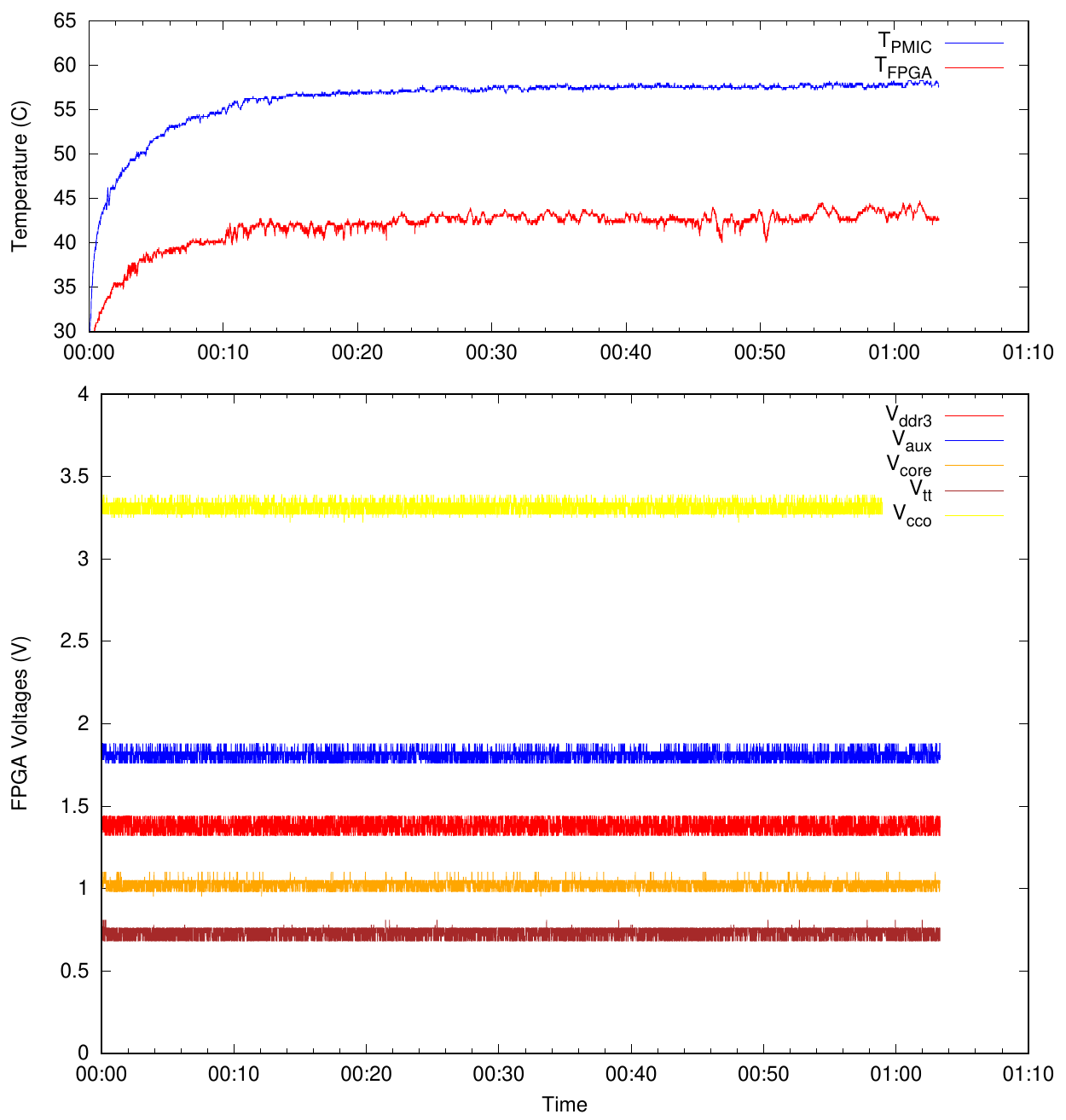}}
\caption{Voltage and temperature sensor reading in an environment without radiation. Temperature and voltage values operate inside certain bounds, and there are few deviations from the mean.}
\label{normal_behaviour}
\end{figure}

\begin{figure}[htb!]
\centering{\includegraphics[width=1\columnwidth]{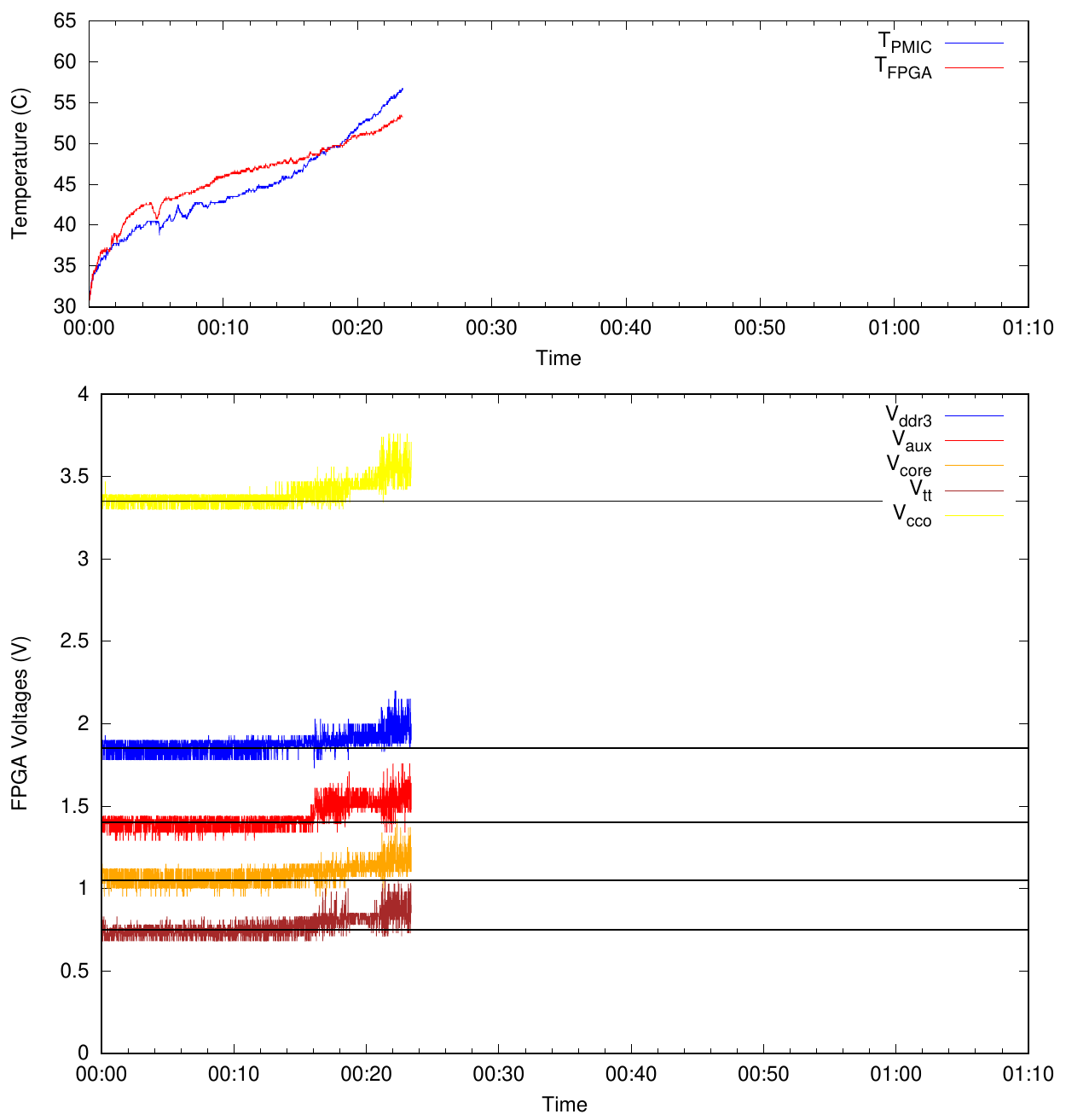}}
\caption{Voltage and temperature sensor reading on an environment with $\gamma$ radiation. Temperature and voltage values operate outside bounds compared to a mean without radiation.}
\label{radiation_behaviour}
\end{figure}

Figure \ref{normal_behaviour} shows the measurements of the Minized board under normal operation. We monitor the power controlling IC and the FPGA temperature ($T_{PMIC}$ and $T_{FPGA}$ respectively) and five voltages ($V_{ddr3}$, $V_{aux}$, $V_{core}$, $V_{tt}$ and $V_{cco}$) supplied to the FPGA. 
One can see a normal fluctuation in the temperature and voltages within some bounds. On the other hand, Figure \ref{radiation_behaviour} shows the same type of board under gamma radiation, where the black lines on voltages are the average of the initial readings. The temperature starts to rise faster, and voltages operate outside their normal bounds and eventually, the board stops working. These two simple examples clearly show that the board is not behaving normally under radiation, as expected. 
Later sections try to correlate these behaviours to predict when the board will stop working. Furthermore, we would use state-of-the-art machine learning techniques to evaluate whether it is possible to indicate when a board will fail based on voltages and temperature sensor readings.

\section{Understanding the statistical effect of gamma radiation on Minized boards}
\label{statistical_effects}


To examine the effects of different levels of $\gamma$ radiation on our boards, we conducted a statistical analysis using measurements for boards that stopped working after being exposed to $\gamma$ radiation and for boards used in a non-radiation environment. To test whether any significant differences occurred in these measurements, we conducted one-way Multivariate Analyses of Variance (MANOVA). The first MANOVA test compared the $\gamma$ radiation effects on the boards (i.e. 0: the measurements collected from boards working under $\gamma$ radiation; 1: measurements obtained from boards on a non-radiation site) as independent variable, and sensor measurements, two temperatures ($T_{PMIC}$ and $T_{FPGA}$) and five voltages($V_{ddr3}$, $V_{aux}$, $V_{core}$, $V_{tt}$ and $V_{cco}$) as dependent variables. The second MANOVA test involved different $\gamma$ radiation levels  (to which boards were exposed) as an independent variable (i.e. $\gamma$ radiation levels detailed in Table~\ref{table_scenarios}) and temperature and voltage  as dependent variables.

The statistical results are presented in Table \ref{tab:anova_results1} and Table \ref{tab:anova_results2}. The partial eta squared ($\eta^2$) represents the effect size, determining how much the relationship will affect the values. On the other hand, the F-value is the test statistic used to determine how much one variable is associated with the response. The factors and interaction effects were analyzed with one-way analysis using the partial eta squared index of effect size. The Bonferroni procedure was used here. The definitions in \cite{Cohen1977} have been adopted to discuss the effect sizes: small effect size ($\eta^2 \leq .01$), medium effect size ($.01 \leq \eta^2 \leq .06$) and large effect size ($.06 \leq \eta^2 \leq .14$). The MANOVA levels of significance are reported using the F-statistics and probability $p$. A risk of $\alpha$ of .05 was used in all statistical tests. 

\begin{table}[htb!]
\renewcommand{\arraystretch}{1.3}
\caption{Results of One-Way Multivariate Analyses of Variance to discover the sensory observation difference between functioning and non-functioning sensor boards used in radiation and non-radiation environments. $\eta^2$ is the partial eta squared measure of effect size. $^\star p < .05, ^{\star\star}p < .01, ^{\star\star\star}p < .001$. The table demonstrates the statistical effect of the main factor. The error degrees of freedom was the same for each dependent variable. }
\label{tab:anova_results1}
\centering
\begin{tabular}{|l|l|crc|}\hline
Source & Dependent Variable &$df$ & $F$ & $\eta^{2}$  \\
\hline
\begin{tabular}[c]{@{}c@{}}$\gamma$ Radiation \\ Functioning\end{tabular} & $T_{PMIC}$ &$1$ & $33051.493^{\star \star \star}$ &$.214$ \\
& $T_{FPGA}$ &$1 $ & $18569.785^{\star \star \star}$ &$.133$ \\
& $V_{aux}$ &$1$ & $1040594.28^{\star \star \star}$ &$.896$ \\
& $V_{ddr3}$ &$1$ & $463097.020^{\star \star \star}$ &$.793$ \\
& $V_{core}$ &$1$ & $30362.611^{\star \star \star}$ &$.200$ \\
& $V_{tt}$ &$1$ & $19258.336^{\star \star \star}$ &$.137$ \\
& $V_{cco}$ &$1$ & $87822.015^{\star \star \star}$ &$.420$ \\

\hline
Error && \multicolumn{3}{l|}{121226}  \\
\hline
\end{tabular}
\end{table}

\begin{table}[htb!]
\renewcommand{\arraystretch}{1.3}
\caption{Results of One-Way Multivariate Analyses of Variance to discover the sensory observation difference for sensor boards at different gamma radiation levels. $^\star p < .05, ^{\star\star}p < .01, ^{\star\star\star}p < .001$. The table demonstrates the statistical effect of the main factor. The error degrees of freedom was the same for each dependent variable. }
\label{tab:anova_results2}
\centering
\begin{tabular}{|l|l|crc|}\hline
Source & Dependent Variable &$df$ & $F$ & $\eta^{2}$  \\
\hline

$\gamma$ Radiation & $T_{PMIC}$ &$6$ & $9721.465^{\star \star \star}$ &$.325$ \\
& $T_{FPGA}$ &$6$ & $4066.381^{\star \star \star}$ &$.168$ \\
& $V_{aux}$ &$6$ & $197999.152^{\star \star \star}$ &$.907$ \\
& $V_{ddr3}$ &$6$ & $94615.951^{\star \star \star}$ &$.824$ \\
& $V_{core}$ &$6$ & $6617.718^{\star \star \star}$ &$.247$ \\
& $V_{tt}$ &$6$ & $4006.448^{\star \star \star}$ &$.165$ \\
& $V_{cco}$ &$6$ & $148.701^{\star \star \star}$ &$.577$ \\

\hline
Error && \multicolumn{3}{l|}{121221}  \\
\hline
\end{tabular}
\end{table}

\subsection{Comparison of boards behaviour on radiation and non-radiation sites}

There was a highly significant effect of the functioning ($Functioning$) of the boards deployed in radiation site when compared to a non-radiation sites in Table \ref{tab:anova_results1}. Overall, there was highly significant effect on the functioning of the sensor boards with very large effect sizes when the measurements collected from radiation and non-radiation sites: $F(1, 121226)=33051.493, p<.001$, $\eta^2=.214$ for temperature 1; $F(1, 121226)=18569.785, p<.001$, $\eta^2=.133$ for temperature 2; $F(1, 121226)=1040594.28, p<.001$, $\eta^2=.896$ for voltage 1; $F(1, 121226)=463097.020, p<.001$, $\eta^2=.793$ for voltage 2; ; $F(1, 121226)=30362.611, p<.001$, $\eta^2=.200$ for voltage 3; ; $F(1, 121226)=19258.336, p<.001$, $\eta^2=.137$ for voltage 4; ; $F(1, 121226)=87822.015, p<.001$, $\eta^2=.420$ for voltage 5. 
 
Our results indicate a significant difference in the obtained measurements between the radiation and non-radiation sites.
This was an expected result, but this is the first work reporting these findings to the best of the authors’ knowledge. It is important to emphasize that since it corroborates the underlying assumption of the proposed analysis of this paper.


\subsection{The effect of different $\gamma$ radiation levels}

There was a highly significant effect of $\gamma$ radiation levels ($\gamma$ Radiation) with a very large effect sizes on the sensor board measurements in Table \ref{tab:anova_results2}: $F(6, 121221)=9721.465, p<.001$, $\eta^2=.325$ for temperature 1; $F(6, 121221)=4066.381, p<.001$, $\eta^2=.168$ for temperature 2; $F(6, 121221)=197999.152, p<.001$, $\eta^2=.907$ for voltage 1; $F(6, 121226)=94615.951, p<.001$, $\eta^2=.824$ for voltage 2; ; $F(6, 121221)=6617.718, p<.001$, $\eta^2=.247$ for voltage 3; ; $F(6, 121221)=4006.448, p<.001$, $\eta^2=.165$ for voltage 4; ; $F(6, 121221)=27547.584, p<.001$, $\eta^2=.577$ for voltage 5.

The posthoc analyses showed highly significant differences between most sensors when exposed to different $\gamma$ radiation levels $p<.001$. However, there were no significant differences in the following interactions. For temperature 1 measurements, there was no significant difference between 2469 and 7707 $\gamma$ radiation levels. For temperature 2, there was no significant difference between 0 and 5871 $\gamma$ radiation levels, and there was only a significant difference between 7707 and 16966 $\gamma$ radiation levels $p<.05$. For voltage 1, there was no significant difference between 2469 and 5137 and 5871 $\gamma$ radiation levels. Similar to temperature 2 results, there was only a significant difference between 7707 and 16966 $\gamma$ radiation levels for voltage 1 measurements. For voltage 2 measurements, there was a significant difference between 7707 and 16966 $\gamma$ radiation levels ($p<.05$). For voltage 3 measurements, there was no significant difference between 1209 and 7707 $\gamma$ radiation levels and between 5137 and 5871 $\gamma$ radiation levels. For voltage 4 measurements, there was no significant difference between 1209 and 7707 and 16966 $\gamma$ radiation levels. For voltage 5 measurements, there was no significant difference between 2469 and 5971 $\gamma$ radiation levels and between 7707 and 16966 $\gamma$ radiation levels.

These results indicate that it is impossible to correlate a given voltage to a radiation level and create a relationship between them. Different sensor inputs would have different weights depending on the radiation rate level. For this reason, more elaborate approaches, such as machine learning algorithms, would have better results since there is a tuning of inputs through experience, in this case, the historical readings of temperature and voltage.


\section{Anomaly Detection with Machine Learning Models}
\label{ML_techniques}

Typically, anomalous data in this study are connected to problems or rare events such as abnormal temperatures or voltages or malfunctioning components. This connection may imply which data points can be considered anomalies to identify these events that are typically useful for predicting the early failure of the system. This section explores three machine learning models: 1) Elliptical Envelope, 2) Local Outlier Factor, 3) One-Class Support Vector Machine as our anomaly detectors. 

\subsection{Machine Learning Models}
\subsubsection{Elliptical Envelope}
Elliptical Envelope is a Gaussian distribution-based method that forms the key data parameters into an underlying multivariate Gaussian distribution expression. In short, it attempts to identify a boundary ellipse that covers most of the data. Therefore, any data not within the ellipse can be classified as an anomaly. The FAST-minimum covariance determinant is used to estimate the size of the ellipse, which selects non-overlapping samples of data and computes the mean $u$ and covariance matrix $C$. Therefore, Mahalanobis distance $d_{MH}$ for the input data vector $x$ can be calculated using the following equation, and the data are then ordered ascendingly by $d_{MH}$ \cite{hoyle15}.

\[{d_{MH}} = \sqrt {{{(x - \mu )}^T}{C^{ - 1}}(x - \mu )} \]

\subsubsection{Local Outlier Factor}
Local Outlier Factor is one of the Nearest Neighbour based methods for anomaly detection. In general, normal data are usually grouped in a neighbourhood that seems dense compared to the abnormal data, which are far from their close neighbours. To quantify this neighbourhood, these types of approaches typically use distance-based or density-based methods, where both ways require a similarity or a distance calculation to determine whether the data are on the degree of abnormality or not. We use the Local Outlier Factor (LOF) abnormal detector in this study \cite{alshawabkeh10}.



\subsubsection{One-Class Support Vector Machine}
One-Class Support Vector Machine (OCSVM) \cite{heller03} is a classification-based anomaly detection method. Depending on the availability of labels, it can be divided into one-class and multi-class classification models. This approach is similar to all other supervised learning techniques, has two phases: 1) Training phase and 2) Testing phase.
In the training phase, the classifier is trained using the labelled data, and then the data are classified as normal or abnormal using the trained model in the testing phase.
In OCSVM, the classification rule for the linear decision boundary is given as follows:
%
\[f(x) = \left\langle {\vec{w},\vec{x}} \right\rangle + b\]%
where $\vec{w}$ and $b$ are the normal vector and bias, respectively. The algorithm is trying to find the rule $f$ within the maximal geometric margin and then assign a label to a test example $\vec{x}$. For example, if $f(x)>0$, the label of $\vec{x}$ will be marked as normal; otherwise, it is labelled anomaly. This optimization problem can be solved by
\[\mathop {\min }\limits_\alpha \frac{1}{2}\sum\limits_{ij} {{\alpha _i}{\alpha _j}K({x_{i,}}{x_j})} \] %
where $\alpha_i$ is the $i^{th}$ weight, $0\le\alpha_i\le\frac{1}{vl}$ and $\sum\limits_{i}\alpha_i=1$. $v$ is a variable to control the maximizing the distance between the origin and the number of data points contained in the boundary. $l$ is the number of points in the training dataset. $K(x_i,x_j)$ is the kernel function, and it is given as follows.
\[K(x,y) = \left\langle {\phi (x),\phi (y)} \right\rangle \] %
where $\phi$ maps the training vectors from input space $X$ to a high dimensional feature space.

A series of mathematical functions, known as the kernel, is used by SVM algorithms. The kernel function takes data as input and translates it into the appropriate form. Different SVM algorithms use various kernel function types. The adopted version, OCSVM, uses the Radial Basis Function (RBF) kernel \cite{broomhead1988radial}.

\subsection{Evaluation Metrics: Precision, Recall, F1 score}

This paper employs three well-known metrics for pattern recognition/classification to evaluate the machine learning models: Precision,  Recall and F1 score \cite{Powers11}. Precision is the fraction of relevance among the retrieved instances, while Recall is the fraction of the total amount of relevant instances. All these metrics would signify better results as they approach the value of 1.

In the context of this paper, there are two kinds of data: annotated and not annotated -- data with an anomaly and without it. The employed models should recognize if the data are an anomaly or not. Suppose we have a given dataset with ten anomalies and the remaining data are not an anomaly. A given ML model identifies eight data points as anomalies, of which five are anomalies (true positive) while the rest are not (false positives). The ML model’s precision is 5/8, while the Recall is 5/10. In this example, precision means how valid the results are, while Recall shows how complete the results are.

F1 score is a metric for accuracy. It considers both the precision and the Recall to compute the score. The F1 score is then calculated using the harmonic mean of the Precision and Recall, where an F1 score has its best value at 1 (perfect precision and Recall).

\section{Radiation Experiments}
\label{rad_experiments}

\begin{figure}[h!]
\centering{\includegraphics[clip,trim={0 50cm 0 55cm},width=1\columnwidth]{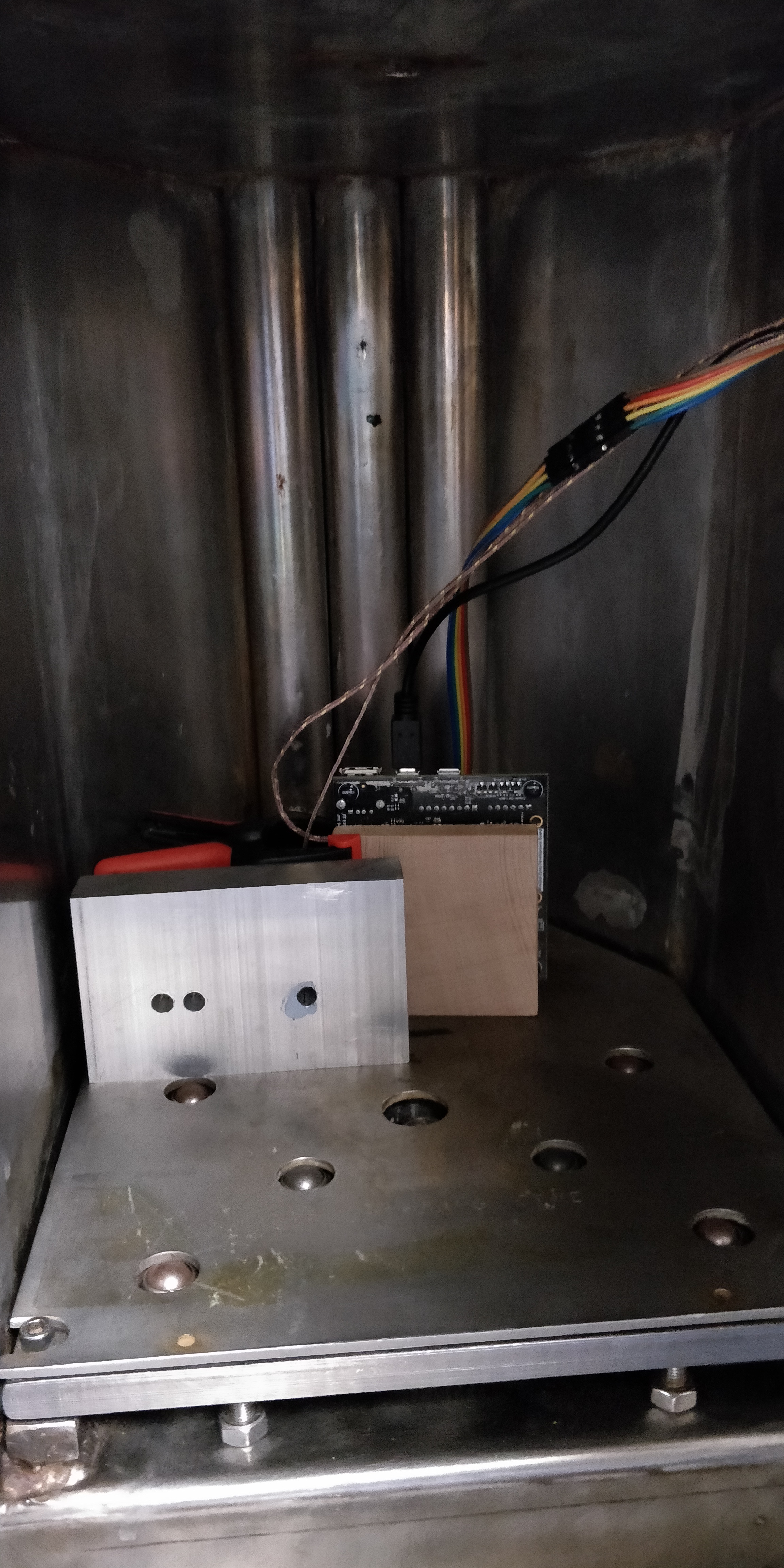}}
\caption{Minized board inside radiation chamber. The three pipes on the back contain three radiation rods that are lifted from the ground when the chamber is closed.}
\label{DCF_rods}
\end{figure}

The experiments took place on the Dalton Cumbrian Facility (DCF) laboratory, where a $\gamma$ radiation source is available \cite{DCF}. \newPart{The source is composed of a Cobalt ($^{60}C$) self-shielded irradiator that can provide absorbed dose rates of up to 20 kGy/h depending on the distance from the source to the DUT. The radiation cavity contains three rods, as shown in Fig.~\ref{DCF_rods}, with different dose rates for each one. Thus, different radiation rates are achieved using various configurations of the available rods, including lead obstacles to absorb the radiation and/or positioning the samples at different distances from the sources.}

\begin{figure*}[ht]
\centering{\includegraphics[width=1\textwidth]{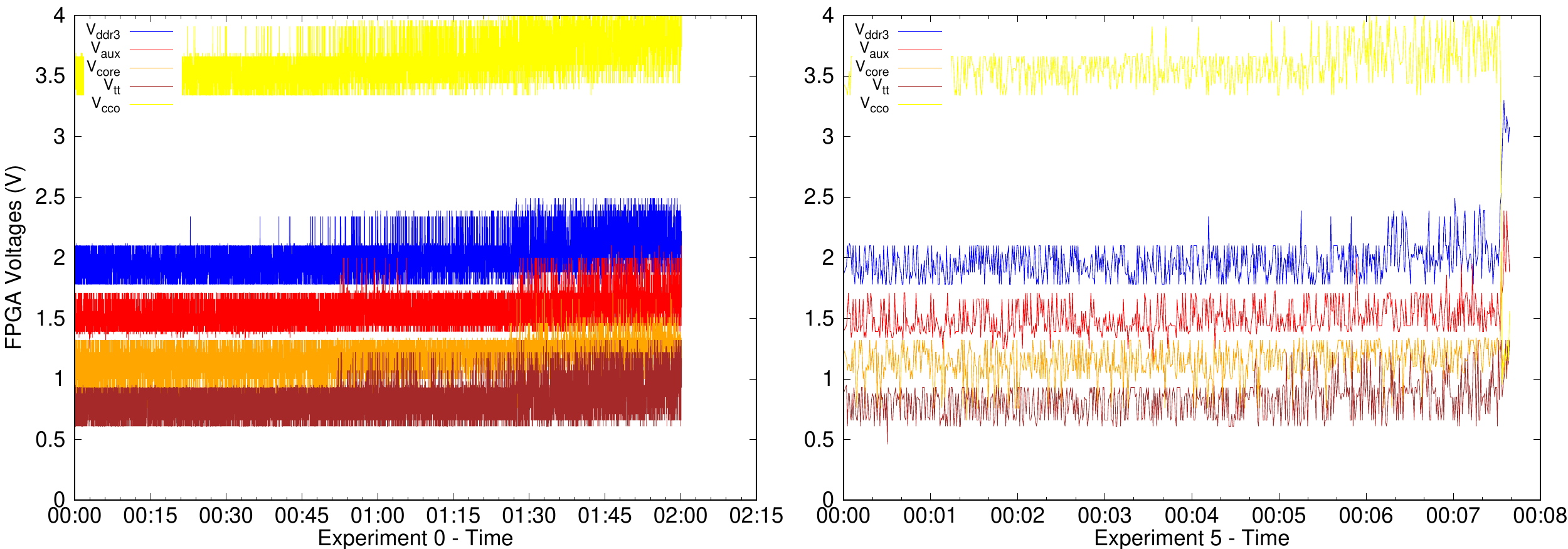}}
\caption{Voltages behaviour of experiments 0 and 5, respectively.}
\label{time_to_fail}
\end{figure*}

\newPart{We employ six different experiments, each one using a separate board (DUT) under radiation -- each board will be referred by its experiment number. Table \ref{table_scenarios} shows each feature of the experiments where different radiation rates were applied to see how the boards would behave. The DUT time is the time between the start of the experiment until the FPGA board stops sending data through the serial connection. This time should be considered the operational time of the DUT. On the other hand, the Monitoring board time is the time from the start of the experiment until the monitored voltages drop to zero, which means that the board itself, including the PMIC, is entirely inoperative. }
In each experiment, the board is under a constant radiation rate and stays the same distance from the rods. Before the actual radiation experiment begins, the radiation rate is measured using a probe removed afterwards. For this reason, the probe is inserted, the radiation is released and measured for one minute, then the experiment is stopped, and the probe is removed. The radiation rate measured during that period is assumed to be constant for the whole experiment.

\begin{table}[htb!]
\renewcommand{\arraystretch}{1.3}
\caption{{Radiation experiment for the DUT}}
\label{table_scenarios}
\centering
\begin{tabular}{|c|c|c|c|}
 \hline

 \textbf{Exp. \#} & \textbf{\begin{tabular}[c]{@{}c@{}}Radiation \\ Rate (Gy/h)\end{tabular}} & \textbf{DUT Time} & \textbf{Monitoring Board Time} \\ [0.5ex]
 \hline

0 & 1209 & 1:56:00 & 2:00:11 \\\hline
1 & 2469 & 0:39:14 & 0:41:31 \\\hline
2 & 5137 & 0:23:16 & 0:27:28 \\\hline
3 & 5871 & 0:19:49 & 0:23:23 \\\hline
4 & 7707 & 0:12:39 & 0:13:39 \\\hline
5 & 16966 & 0:06:58 & 0:07:38 \\

 \hline
\end{tabular}
\end{table}


After the experiments, all the boards stopped working. During the experiments, one can see an increase in the voltage bounds on all experiments. The interesting point is that the deviation from the normal bounds does not indicate that the board will stop working right away. If we take, for example, the two extremes for the radiation rates, i.e. experiments 0 and 5, one took almost two hours while the other took less than ten minutes to stop working. Fig.~\ref{time_to_fail} shows that although experiment 5 exhibits an early change in the bounds of the voltages, the board can take more than one hour to stop working. On the other hand, the board might take a few minutes to stop working with a higher radiation rate after the first values outside the normal bounds are observed. This observation leads to a search for a more elaborate way of predicting when the board will stop working than just watching voltages outside the normal bounds of operation.



\section{Evaluations}
\label{evaluations}

This Section compares the OCSVM method against the Local Outlier Factor and Elliptical Envelope methods. OCSVM copes well with non-linear functions and might be a more suitable approach for this problem.
The section is divided into four subsections. The first one shows how the data was preprocessed and explains the methodology used in the second subsection, where the training set is built. Then, the third subsection compares the OCSVM against the two other methods using the training set. Finally, the last subsection further explores how early we can detect anomalies with OCSVM.

\subsection{Training and Testing Methodology}


This subsection explains the steps to build the training data, later used to compare different approaches. The objective here is to compare how the training data would affect the results rather than comparing different ML algorithms. Hence, we show the comparison using the OCSVM algorithm.

The first step is to remove as least from the collected data, all of which are time-stamped. Unfortunately, a few points show the temperature as zero or undefined (e.g. NaN). Therefore, all data points for that period were removed, even if another sensor showed consistent data. These removals do not interfere with the evaluation since all the data points are time-stamped. 

\begin{figure}[h!]
\centering{\includegraphics[width=1\columnwidth]{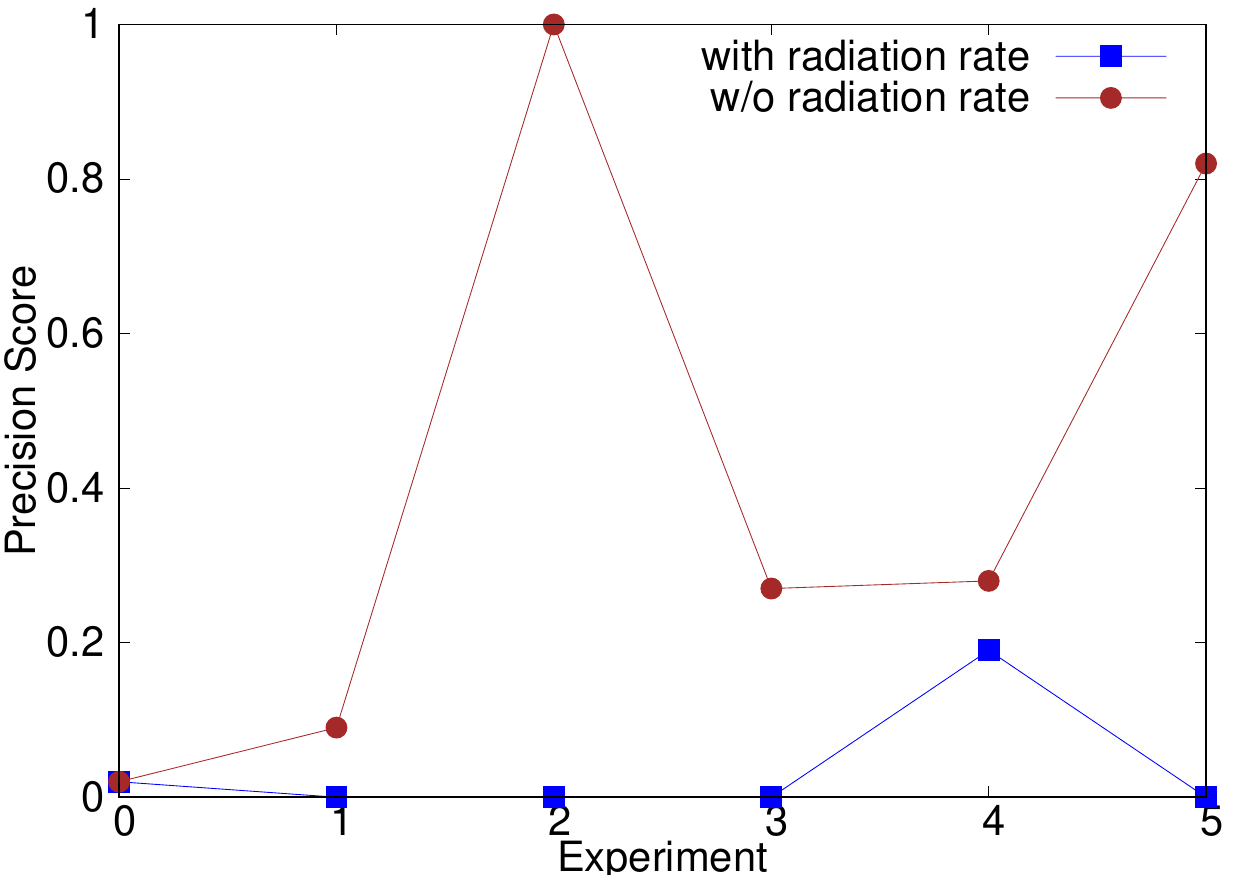}}
\caption{Comparison using different features as input for the training data. Respectively lines show the results using $i$) seven features (five voltages and two temperatures) and $ii$) the same seven features plus the constant radiation rate.}
\label{compare_radiation}
\end{figure}

After the data trimming step, we compare two approaches: one using seven features (five voltages and two temperatures) and one using the same features plus a constant value for the radiation rate -- the rates shown in Table~\ref{table_scenarios}.
Fig.~\ref{compare_radiation} shows the precision of the two models with the six experiments. Using the constant radiation rate as input weakens the results significantly. Therefore, for further experiments, only voltage and temperature values are considered.

\subsection{Exploring the best training and testing strategy in a radiation site} 

At this point, the number of features is defined, we came up with three training and testing strategies to evaluate, compute or organize the measured data. The target is to find the right balance of data points in the training set. The following strategies evaluate trade-offs between a set of experiments employed and the number of data points. Each strategy represents a data processing approach, and it is detailed next as follows:
\begin{itemize}
  \item Strategy 1: Train the model with the first minutes of a given board and then test on the remaining measurements of the same board.
  \item Strategy 2: Train the model with the first minutes of a set of different experiments and test on the remaining measurements of all boards.
  \item Strategy 3: Choose the number of data points that gives the best results in terms of F1 score.
\end{itemize}

\begin{figure}[htb!]
\centering{\includegraphics[width=1\columnwidth]{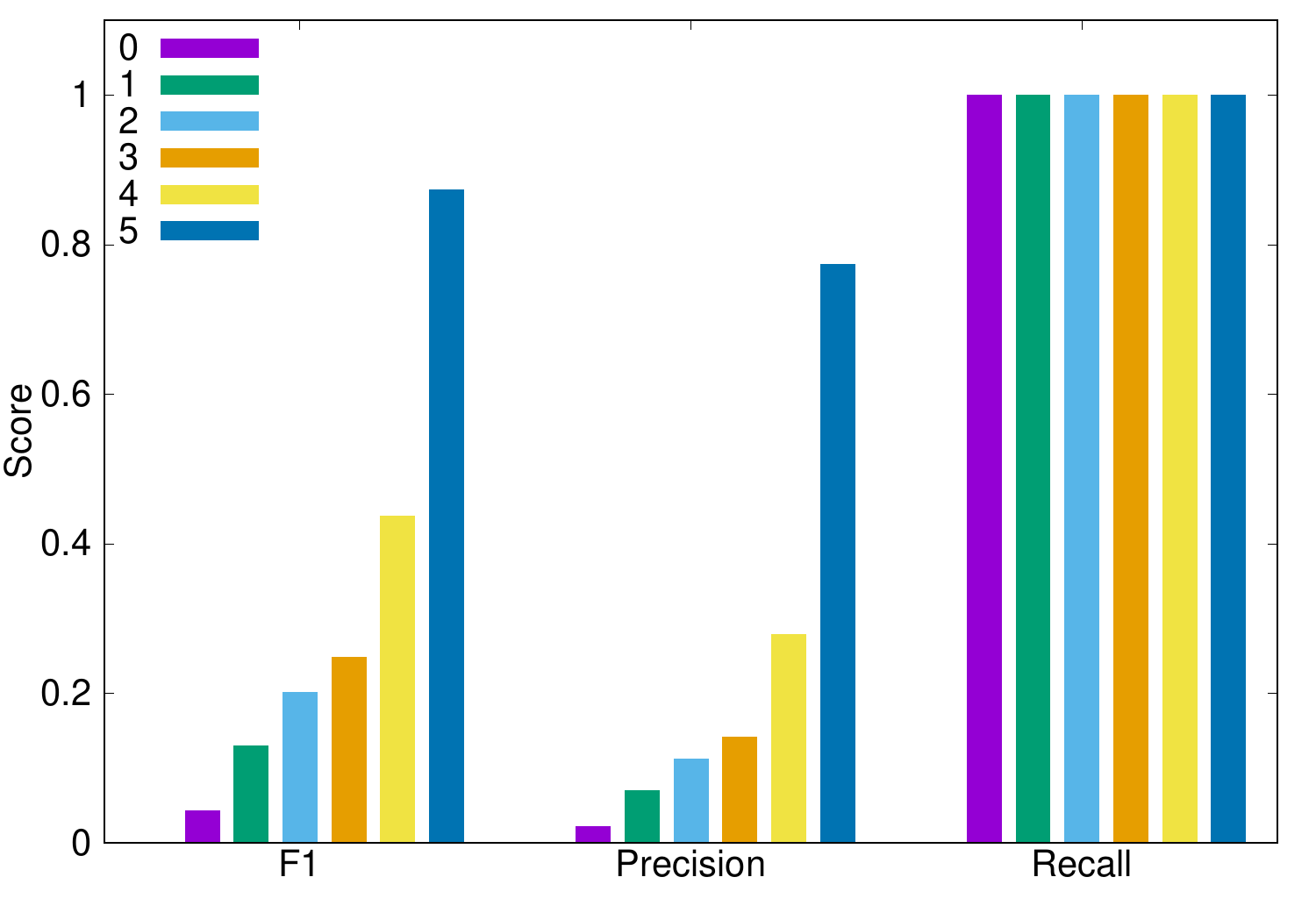}}
\caption{Results using each experiment separately as training data. }
\label{compare_strategy_1}
\end{figure}

The first strategy was employed using the first few minutes of each board and then test on the remaining measurements on the same board. Fig.~\ref{compare_strategy_1} shows the results for this strategy. Even using the same board data and comparing it with the remaining data does not provide good results. Only board from experiment 5 shows good scores for F1 and precision.

\begin{figure}[htb!]
\centering{\includegraphics[width=1\columnwidth]{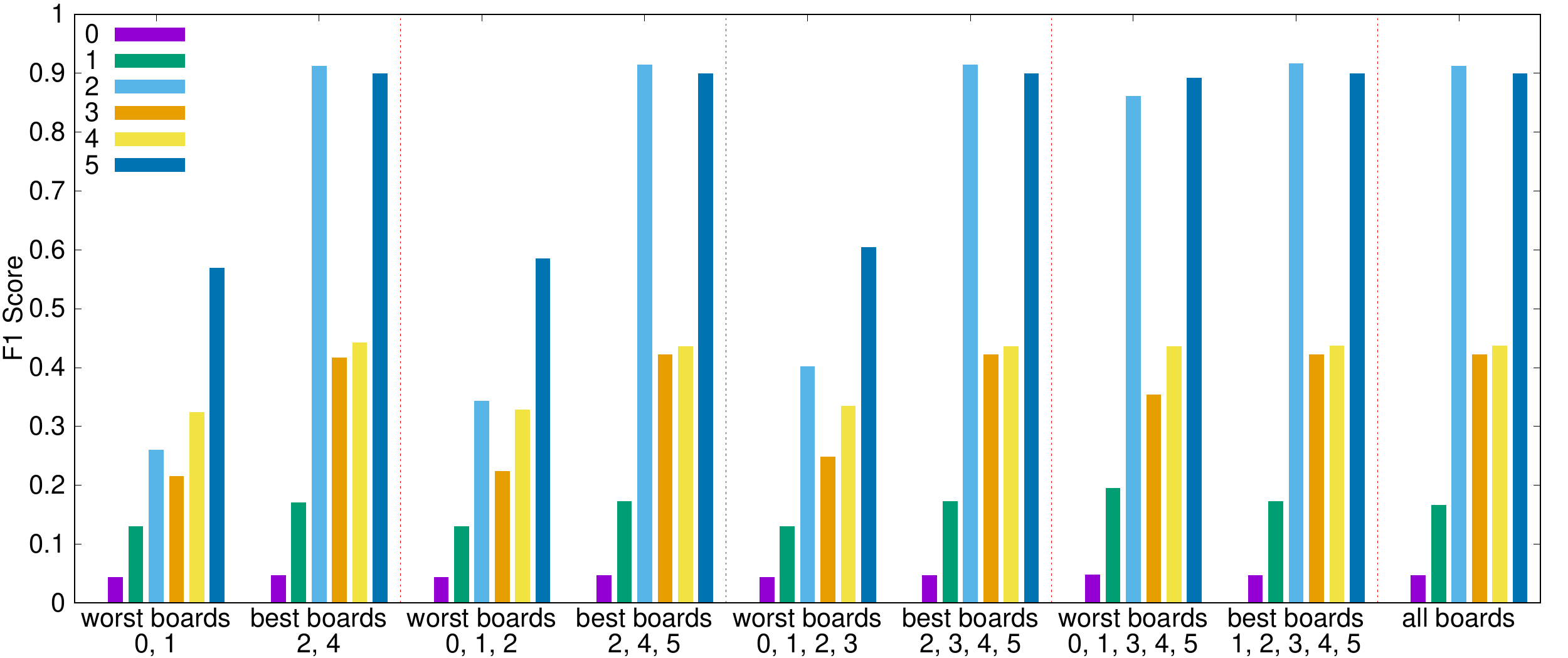}}
\caption{Using sets of 2, 3, 4, 5 and all boards as training data. Figure showing only best and worst results for each set.}
\label{compare_strategy_2}
\end{figure}

\newPart{Because of the weaker results of strategy 1, we decided to include more boards on the training data.
Strategy 2 evaluates different sets of boards as training data to check how many we need to include to get the best results. For this strategy, we trained all combinations sets of two up to all six boards, i.e. sets of two boards \{\{0,1\}, \{0,2\}, ... \{4,5\}\}, sets of three boards \{\{0,1,2\}, \{0,1,3\}, ... \{3,4,5\}\} and so on. Later we compared the F1 score for each board separately.} Results shown in Fig.~\ref{compare_strategy_2} include the worst and best set of boards only for the sake of space. One can assume that we get better results as we feed more information to the model (different boards). For example, the worst results for sets with two, three and four boards are much lower than results using five or all six boards.

\begin{figure}[htb!]
\centering{\includegraphics[width=1\columnwidth]{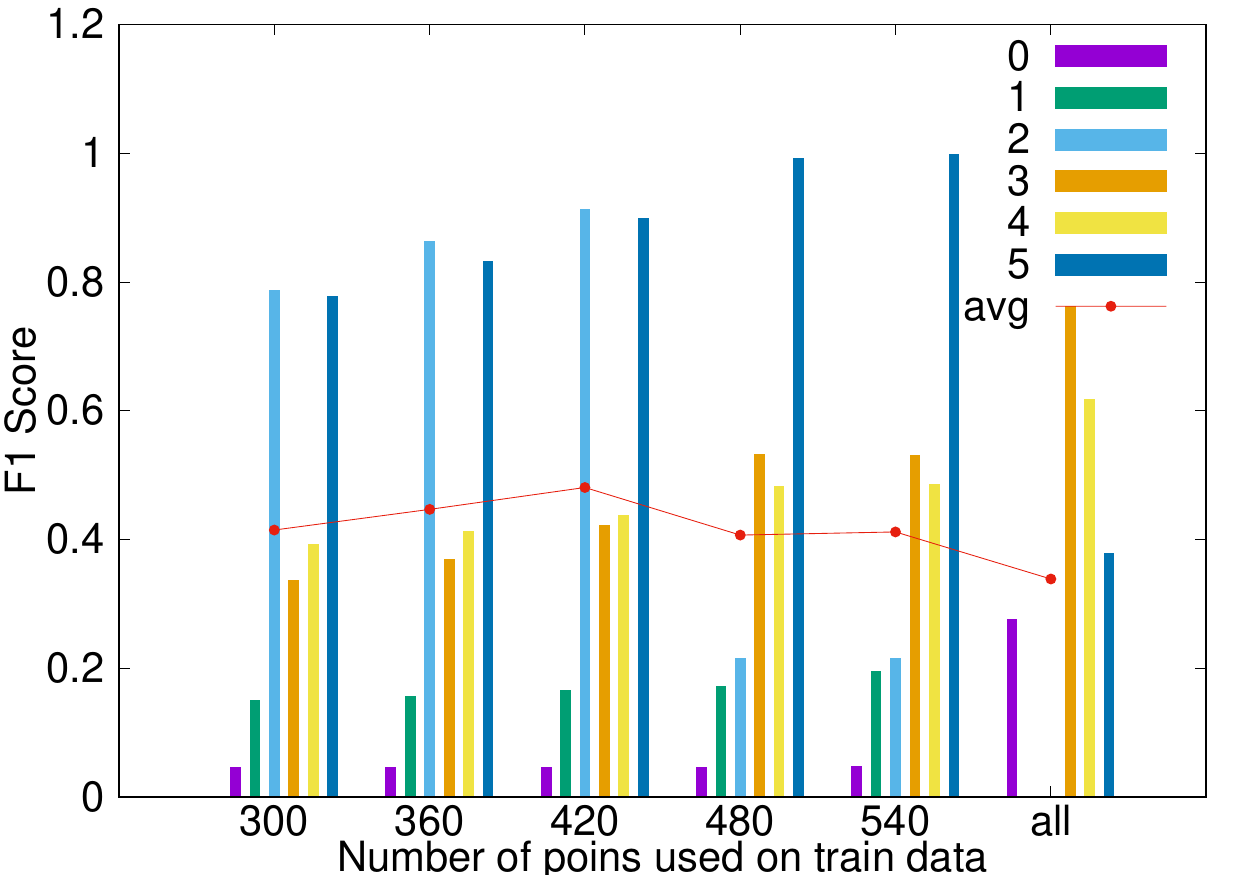}}
\caption{Comparison of training data for input data. Using the first 420 data points from all experiments show better results. 420 data points represent roughly seven minutes of data.}
\label{compare_training_time}
\end{figure}

On the third strategy, we evaluate the effects of the amount of training data fed into the model. Fig.~\ref{compare_training_time} shows the comparison of using the first 300, 360, 420, 480, 520 and all data points as input for the model. Interestingly, as we feed more data to the training set, it does not mean we would get better results. Adding more than 420 points does not increase the F1 score, and as it shows better average results, we keep this number of data points as a training set for all experiments.

\subsection{Comparison of OCSVM with other anomaly detection techniques}
\label{results}


This subsection compares three state-of-the-art machine learning techniques: $i$) Elliptical Envelope, $ii$) Local Outlier Factor and $iii$) One Class Support Vector Machine (OCSVM). These techniques are trained using the proposed dataset, discussed previously, as inputs.

\begin{table}[htb!]
\renewcommand{\arraystretch}{1.3}
\caption{Window times where voltages exhibit values outside usual bounds for each experiment.}
\label{visual_anomalies}
\centering
\begin{tabular}{|c|c|}
 \hline
 \textbf{Exp. \#} & \textbf{Window Time}\\ [0.5ex]
 \hline
0 & 74 min\\\hline
1 & 18 min\\\hline
2 & 07 min\\\hline
3 & 07 min\\\hline
4 & 06 min\\\hline
5 & 03 min\\

 \hline
\end{tabular}
\end{table}

\begin{figure*}
\includegraphics[width=0.33\textwidth]{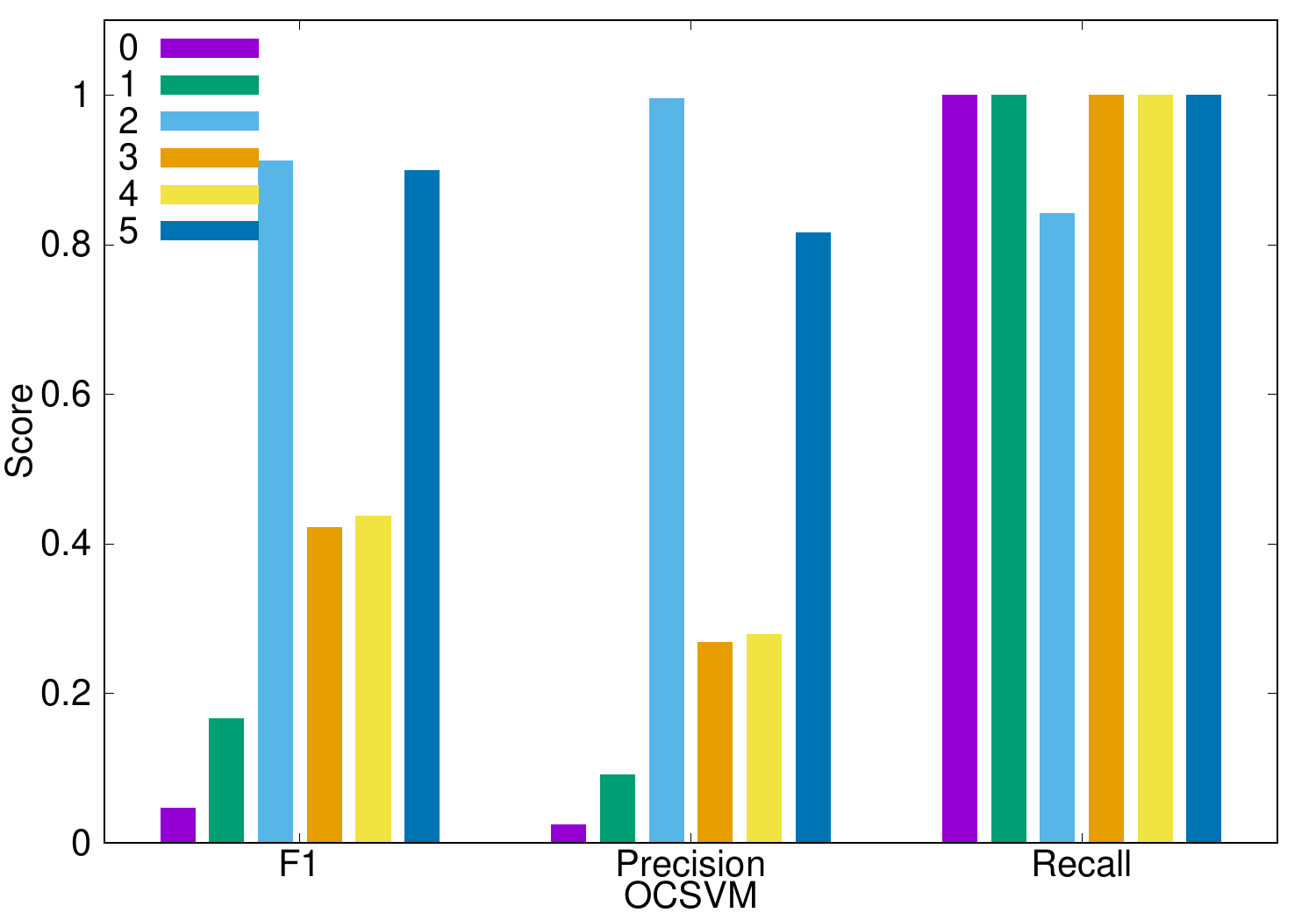}
\includegraphics[width=0.33\textwidth]{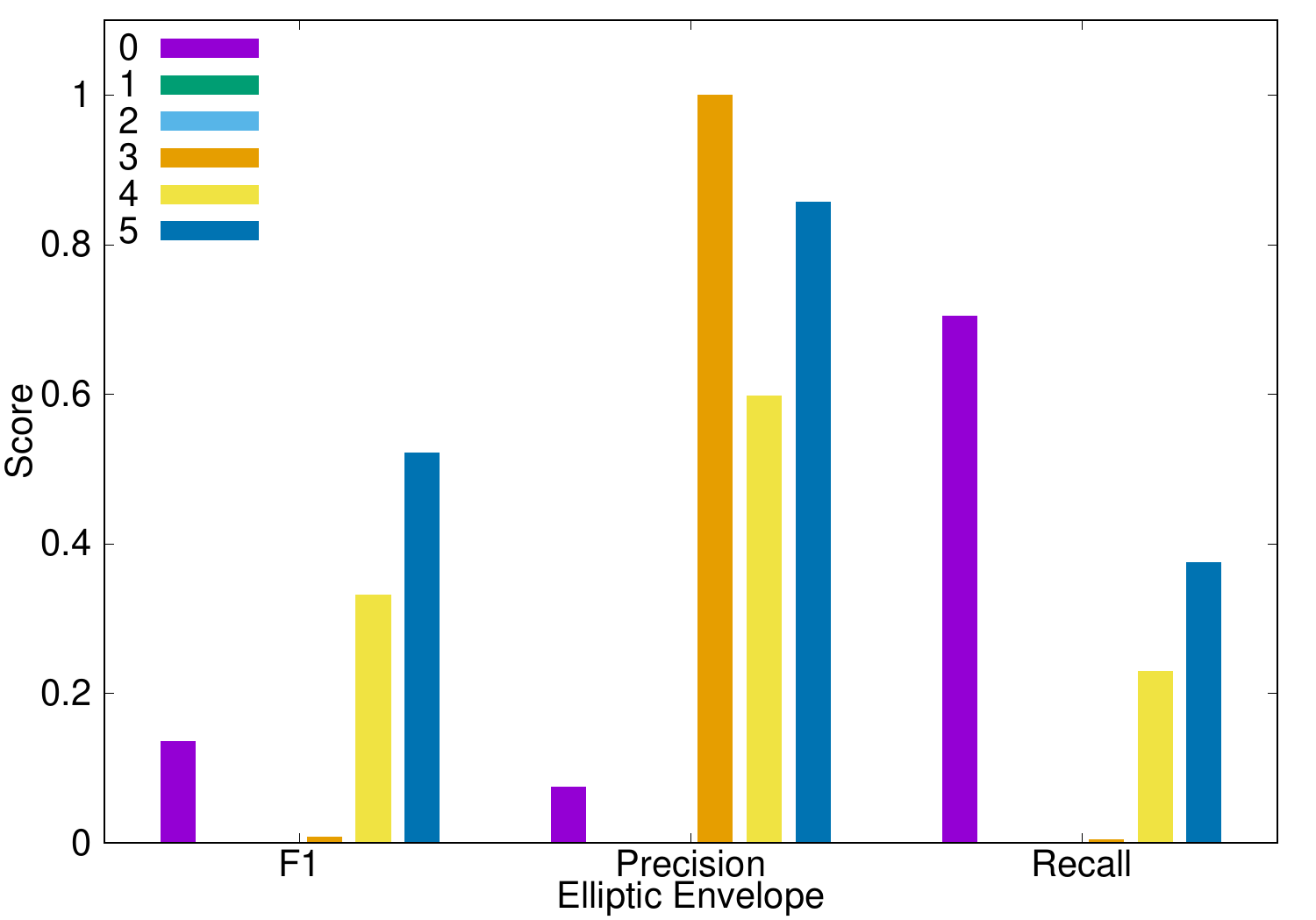}
\includegraphics[width=0.33\textwidth]{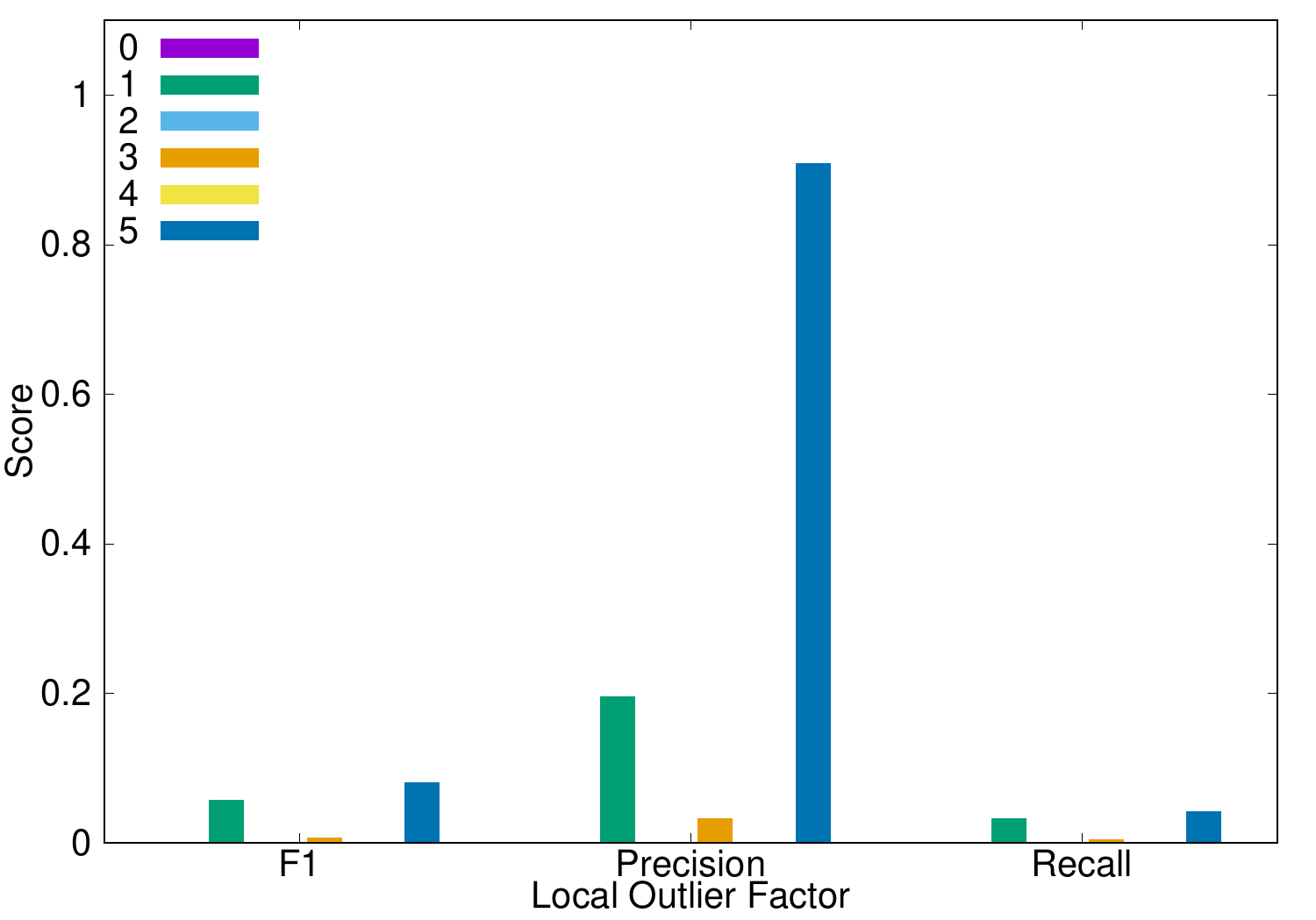}
\caption{Comparison of machine learning models for the dataset.}
\label{compare_ML_approaches}
\end{figure*}

To compare the anomaly detection results, we first need to annotate the data showing where the anomalies happened and then compare with the ML algorithms outputs.
We have calculated the size of the time window where each board exhibits values outside usual bounds during radiation experiments, visually observing the measured voltage (we could not observe such values for temperature). Table ~ shows the size of this window, and associated calculations are displayed for each board in Table~\ref{visual_anomalies}. As can be seen, the minimum window size observed is approximately 3 minutes. That is to say that the last 3 minutes of each experiment is the time where each board exhibits values outside normal bounds. Therefore, after adding a safety margin of 2 minutes, we annotate the last 5 minutes of each experiment as anomalies for use in the training dataset fed into the machine learning algorithms. Five minutes is roughly equivalent to 300 data points; therefore, we annotate this number of points as an anomaly.
Besides the observed window size, 300 data points also give the developer a reasonable amount of time to take action before the board stops working,  e.g., move the computed data to safe storage, to another computing node in the system or even move it away from the radiation environment.
We can also consider a different amount of time as annotation, but it would result in the retraining of the model.
Then finally, we can summarize the training set with the annotation as follows:

\begin{itemize}
  \item Remove all the inconsistent sensor readings;
  \item Collect the first 420 data points from all boards (as justified in Fig.~\ref{compare_training_time});
  \item We annotate the last 300 data points as an anomaly.

\end{itemize}


Using that training set, we then employ a multivariate analysis with seven variables (including five voltage and two temperature values) to feed the ML algorithms. Fig.~\ref{compare_ML_approaches} summarizes the results for Elliptical Envelope, LOF and OCSVM algorithms where the F1, Precision, and Recall scores are shown for each model. OCSVM demonstrates better outcomes for all scores. It is essential to highlight that the Recall score for OCSVM is the one for all experiments except number two, which is a remarkable result, showing an average Recall score of 0.95 and strong evidence that the model can detect the relevant results. The recall score in experiment 2 is not the maximum one but shows a high value (0.842) and will be discussed next.


\newPart{\subsection{Exploring how early OCSVM can detect anomalies}}

\begin{figure*}
\includegraphics[width=0.33\textwidth]{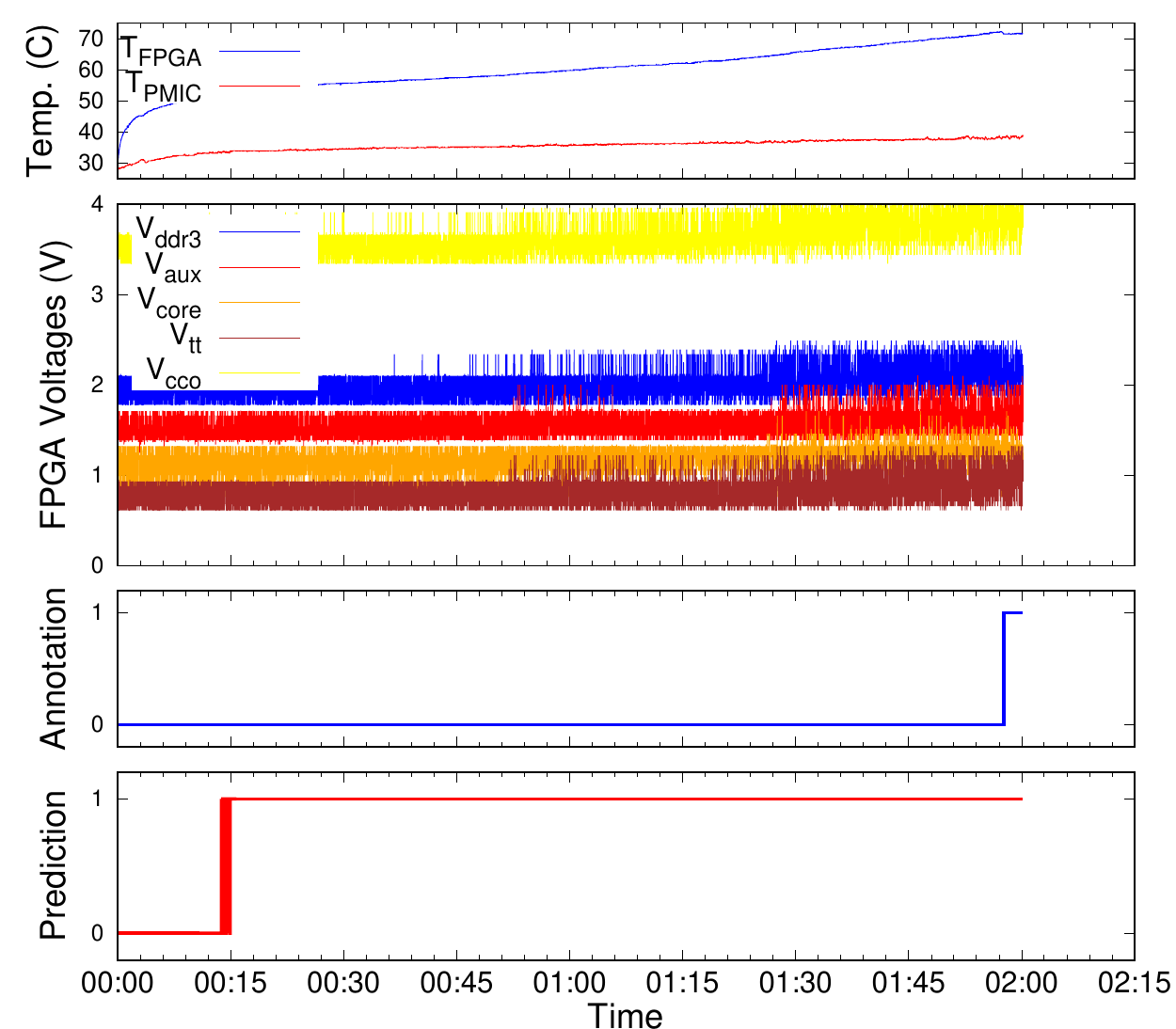}
\includegraphics[width=0.33\textwidth]{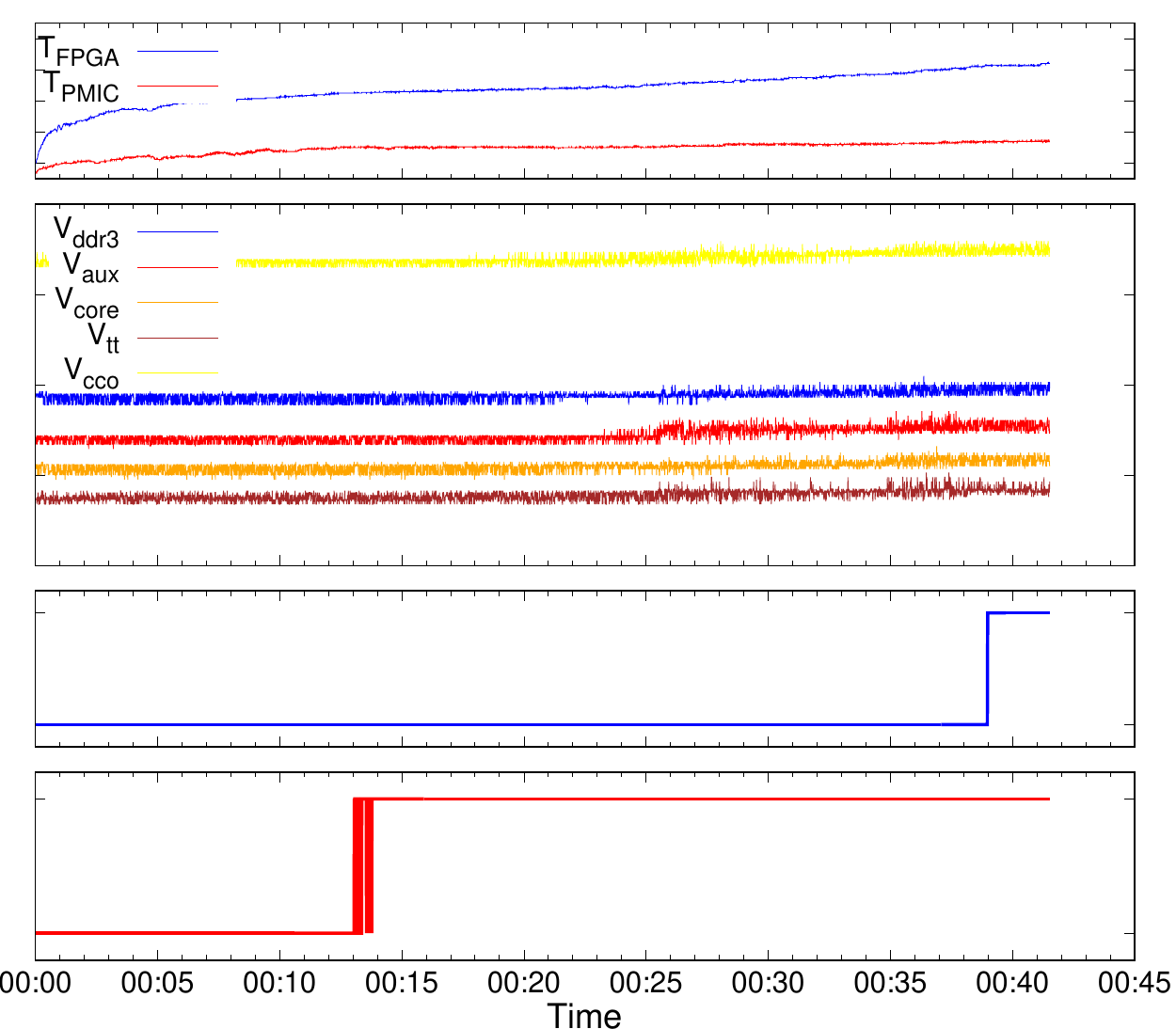}
\includegraphics[width=0.33\textwidth]{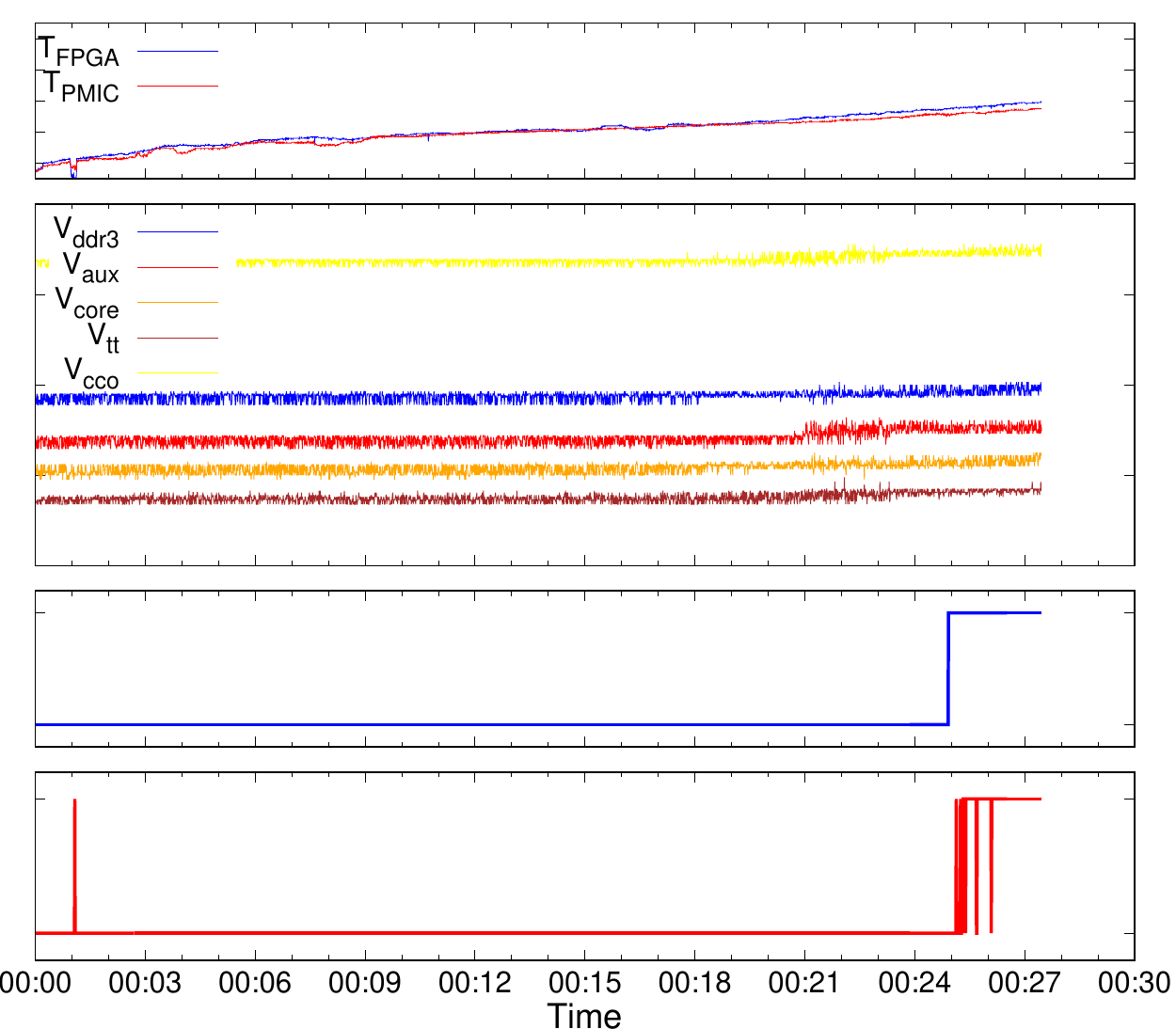}
\begin{tabular}{p{0.13\textwidth} p{0.30\textwidth} p{0.30\textwidth} p{0.30\textwidth}}
  & (a) - Exp. 0 & (b) - Exp. 1 & (c) - Exp. 2 \\
\end{tabular}

\includegraphics[width=0.33\textwidth]{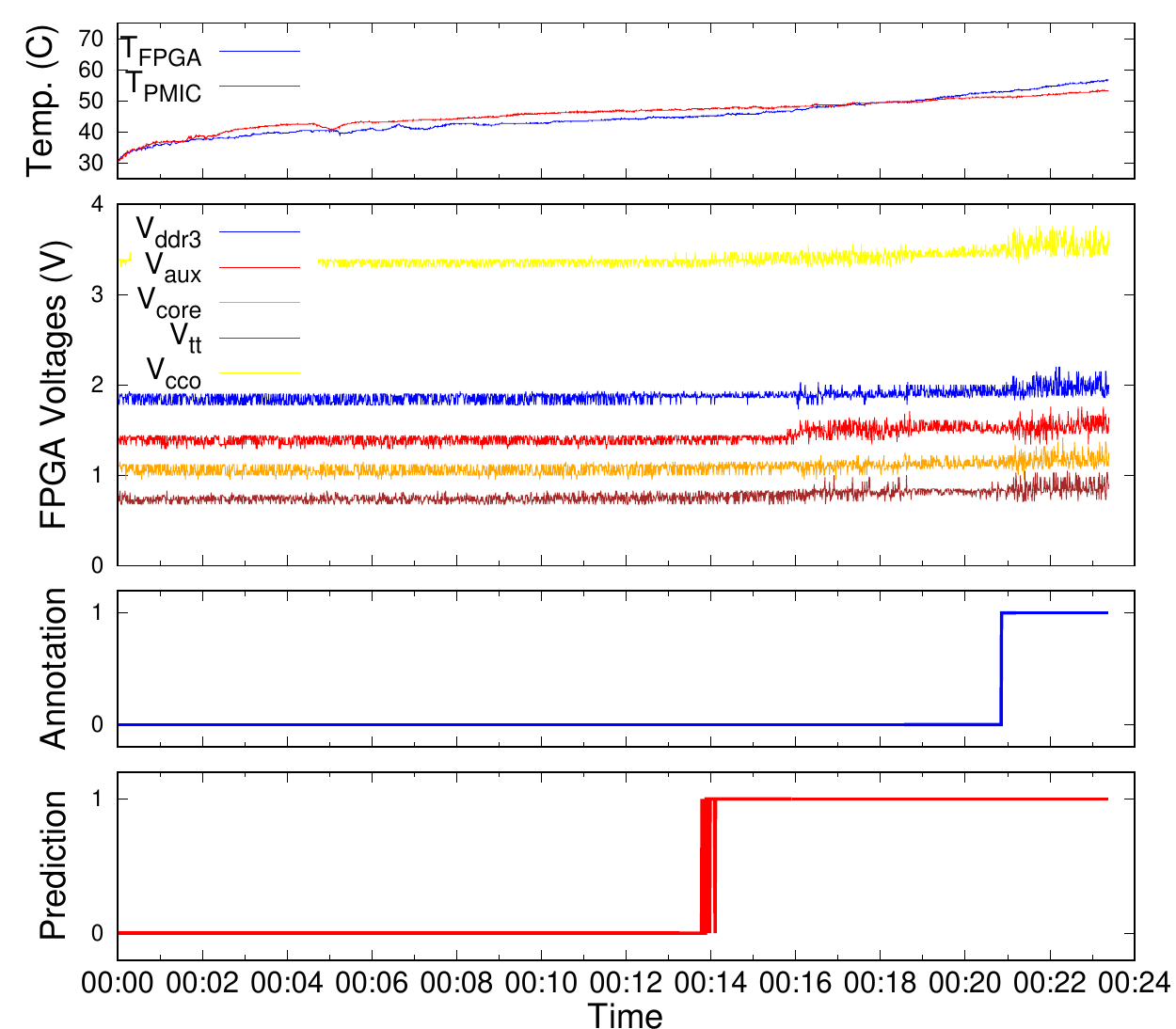}
\includegraphics[width=0.33\textwidth]{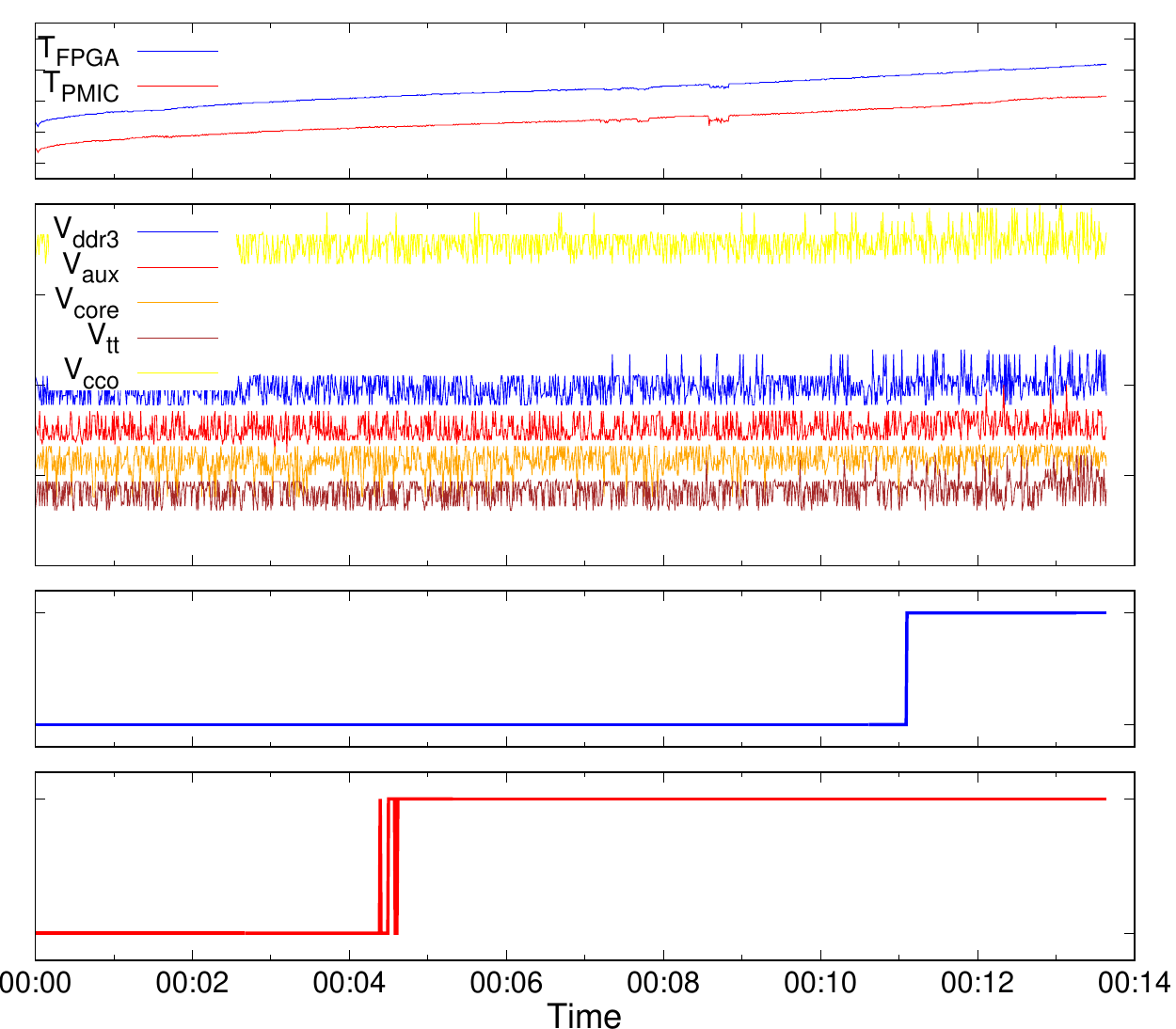}
\includegraphics[width=0.33\textwidth]{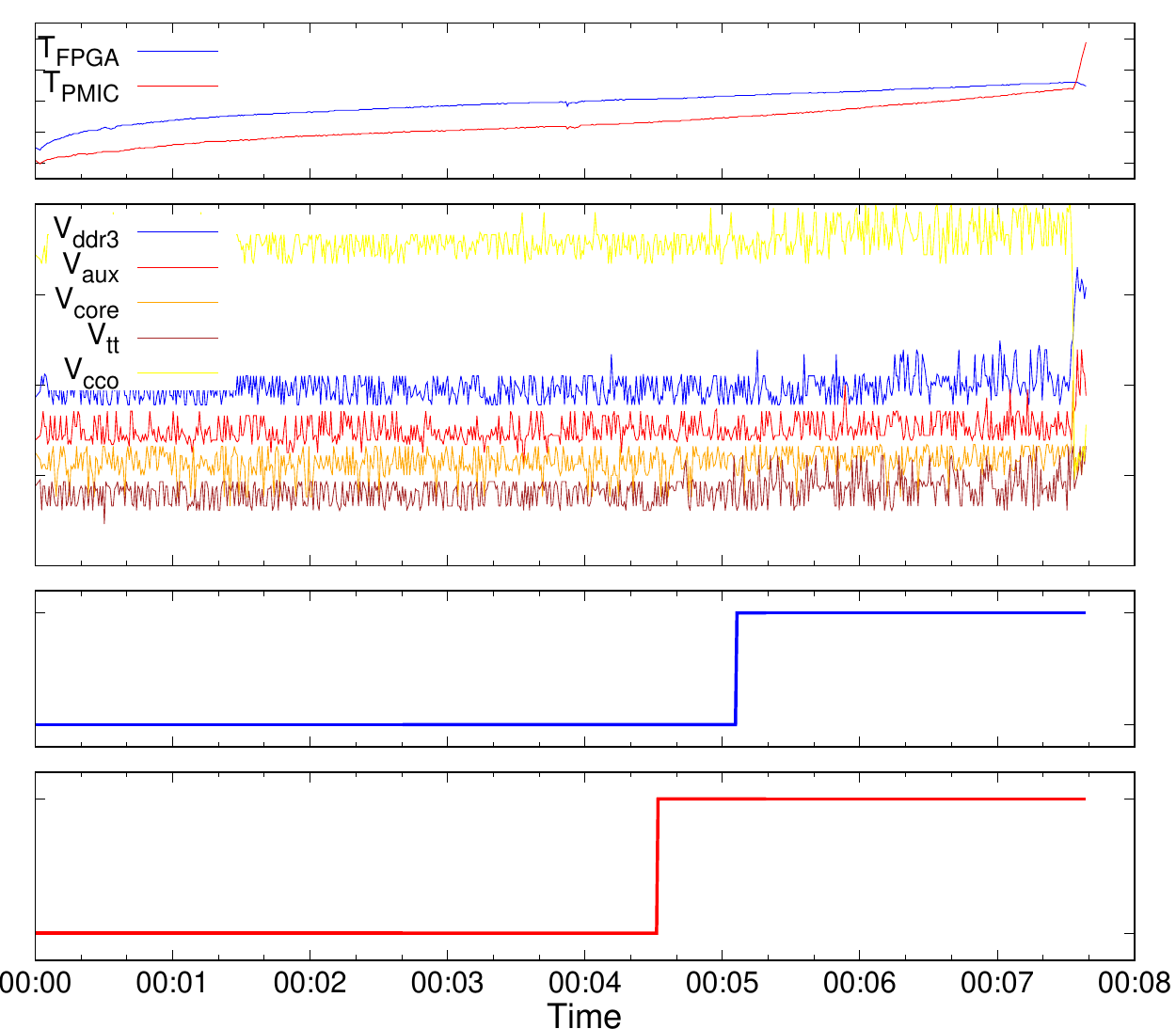}
\begin{tabular}{p{0.13\textwidth} p{0.30\textwidth} p{0.30\textwidth} p{0.30\textwidth}}
  & (d) - Exp. 3 & (e) - Exp. 4 & (f) - Exp. 5 \\
\end{tabular}
\caption{Results for each experiment using One-Class Support Vector Machine to predict the anomaly before the board stops working.}
\label{OCSVM_results}
\end{figure*}

As the OCSVM showed the best F1, Precision and Recall results, this subsection details each experiment separately, as illustrated in Fig.~\ref{OCSVM_results}. Each experiment shows in the rows: (i) the five observed voltages; (ii) two temperatures; (iii) anomaly annotation; and (iv) the OCSVM model output (i.e. 0 represents normal data and 1 an anomaly both for annotation and model output). 

Table~\ref{saving_boards} shows that in all experiments except one, the model marks the abnormal behaviour of the board before the annotation. This is an exceptional result, indicating that the model can be used as a suitable indicator/warning before the board stops working. Note that the ML output for experiment 2 is delayed by only 12 seconds; nevertheless, there still would be enough time for the board to be saved before it is permanently damaged.

\begin{table}[htb!]
\renewcommand{\arraystretch}{1.3}
\caption{Time showing how earlier the OCSVM marks the annotation (around 5 minutes before the board stops working). The model prediction is calculated with (Annotation - Prediction) from Fig.~\ref{OCSVM_results}.}
\label{saving_boards}
\centering
\begin{tabular}{|c|c|c|}
 \hline
 \textbf{Exp. \#} & \textbf{Experiment duration} & \textbf{\begin{tabular}[c]{@{}c@{}}Model prediction \\ before the annotation\end{tabular}} \\ [0.5ex]
 \hline
 
0 & 2:00:11 & 1:43:53 \\\hline
1 & 0:41:31 & 0:25:56 \\\hline
2 & 0:27:28 & -0:00:12 \\\hline
3 & 0:23:23 & 0:07:04 \\\hline
4 & 0:13:39 & 0:06:42 \\\hline
5 & 0:07:38 & 0:00:34 \\
 

 \hline
\end{tabular}
\end{table}

Experiment 0 is the longest in time under radiation mainly because it has the lowest radiation rate among the experiments. The board took almost two hours to stop working, and it is the worst result for the model output. Although the model could point to the anomaly, it has done it earlier, more than one hour before the board stops working.
Experiment 1 has a shorter execution time, and the model was also capable of predicting before the anomaly, i.e. 25 minutes before the annotation.
Experiment 2 has one particular behaviour different from the others. The model output shows one unique point marking as an anomaly in the first minutes. As this was the only point and not a continuous trend, we can disregard this result - one should always observe the trend rather than particular points. The model then marks only points 12 seconds after the anomaly annotation. We still  consider this a good result since a few seconds later would still allow time to take action.

In experiments 3, 4 and 5, the model behaves similarly, marking the anomaly 7, 6 and 0.5 minutes before the annotation, which is a remarkable result. To summarize, all boards would have been saved since the model can point out before a board dies completely, thus allowing a few minutes for the designer to save or transfer the processed data to a safe environment.

It is essential to point out that the model has a trade-off. As we gathered more experiments with high radiation rates, five of the experiments have more than 2000 Gy/h; the model has been trained with more data for these environments. Therefore it also performs better, i.e. predicts near the annotation, on experiments with high radiation rates.
As we feed more information, that is, more experiments in different radiation rates, the model should perform better.



\section{Conclusion and Future Works}
\label{conclusion}

This paper proposed an anomaly detection machine-learning algorithm to predict when a COTS FPGA would stop working due to gamma radiation.
The OCSVM algorithm showed the best results in terms of Recall score and was capable of pointing out 100\% of the anomalies before the board stopped working. Annotating the anomaly before it stops working and giving time to the designer to take actions allowed to detect the anomaly before the board dies, considered an extraordinary achievement since it will enable the designer to use this as an assumption for future works.
This work employed six boards that were inoperable after the experiments. Using more boards with different radiation rates can improve the model results. 

Future works include using the OCSVM algorithm at run-time experiments. The proposed approach  with DUT and a monitoring board can also be modified to execute on a self-contained solution. The Arduino board of the DUT was used to remove the monitoring board and sensors from the radiation environment. In a real-case scenario, this would not be possible. To tackle that approach, the Minized board (DUT) could use the sensor boards directly and execute the OCSVM algorithm itself.


\section*{Acknowledgment}

This work is supported by the U.K. Engineering and Physical Sciences Research Council through grants EP/R02572X/1 and EP/P017487/1.

\bibliographystyle{IEEEtran}
\bibliography{IEEEabrv, Bibliography}

\begin{IEEEbiography}[{\includegraphics[width=1in,height=1.5in,clip,keepaspectratio]{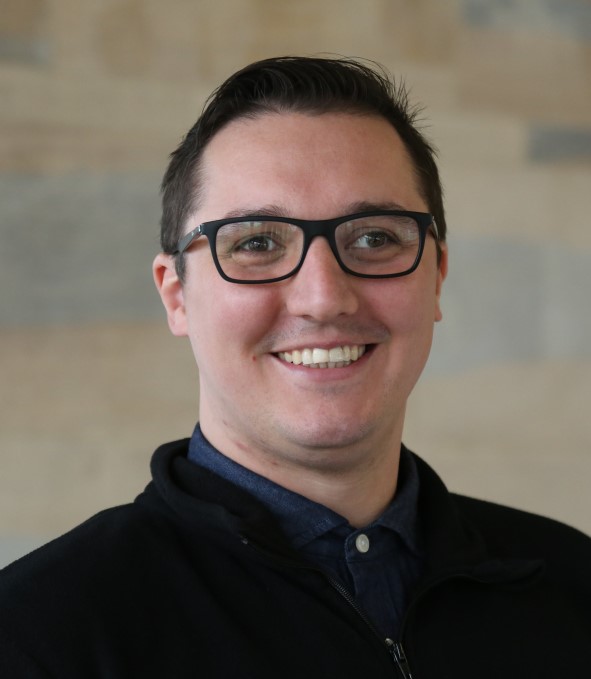}}]{Eduardo Weber Wachter}  received the
B.Eng. degree in computer engineering from the
State University of Rio Grande do Sul, Guaíba,
Brazil, in 2009 and the PhD degree in computer
science from the Pontifical Catholic University
of Rio Grande do Sul, Porto Alegre, Brazil, in
2015. His research interests are Many Cores, NoC,
reconfigurable architectures, Fault Tolerance and
Reliability of such systems. He is currently a Teaching Fellow with the University of Warwick, U.K.
\end{IEEEbiography}

\begin{IEEEbiography}[{\includegraphics[width=1in,height=1.5in,clip,keepaspectratio]{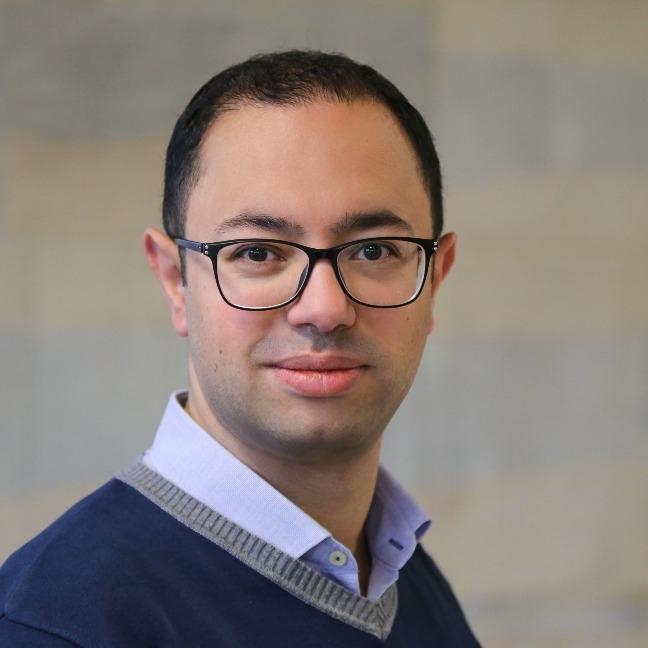}}]{Server Kasapr} (M’10) received the B.Sc.
(Hons.) degree in electrical and electronic engineering
from Middle East Technical University,
Ankara, Turkey,and the M.Sc.(Distinction) and
PhD degrees in electronic engineering from the
System-Level Integration Research Group, University
of Edinburgh, Edinburgh, U.K., in 2006,
2007, and 2010, respectively. His current research interests include reconfigurable dataflow computing for finite-difference time-domain simulations, FPGAhardware design and implementation for digital signal processing applications, and high-performance scientific computing in general. He is currently a Lecturer at Coventry University, U.K.
\end{IEEEbiography}

\begin{IEEEbiography}[{\includegraphics[width=1in,height=1.5in,clip,keepaspectratio]{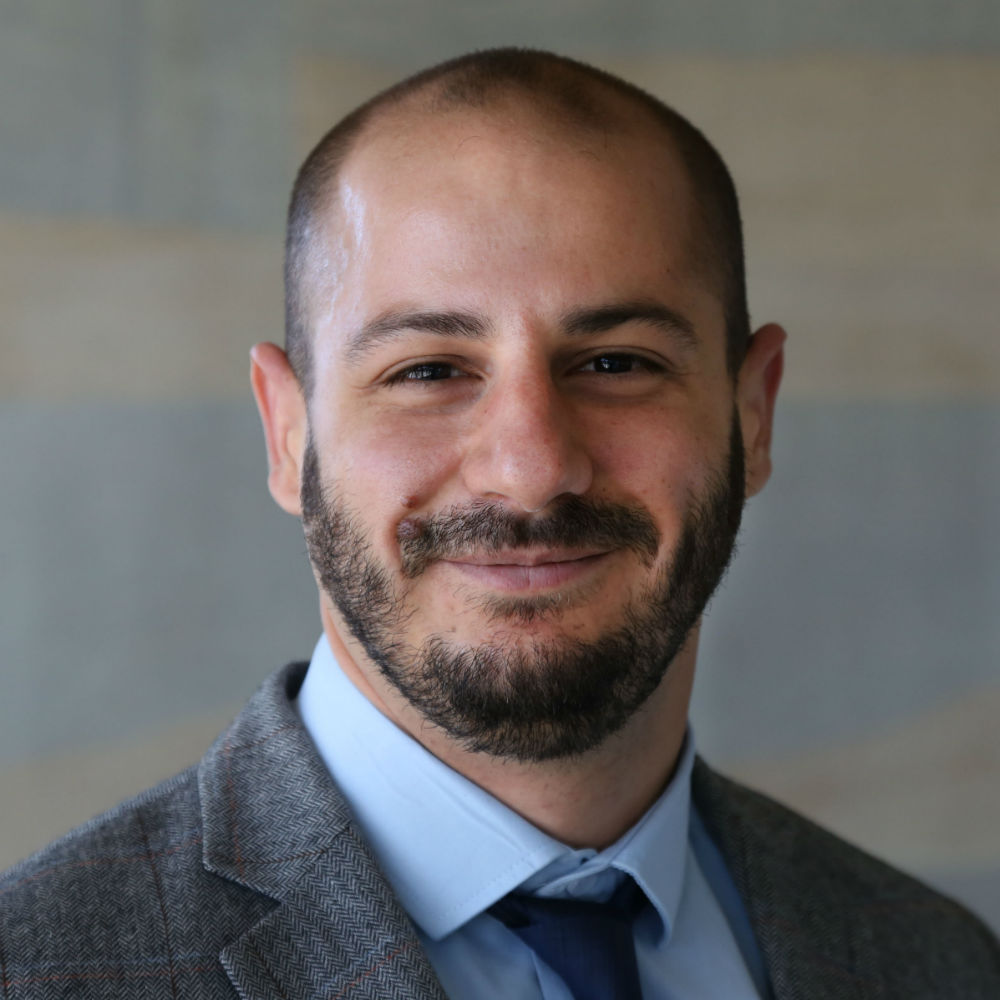}}]{\c{S}efki Kolozali}  received the B.Sc. degree in
computer engineering from Near East University,
Nicosia, Cyprus, in 2005, the M.Sc. degree from the
University of Essex, Colchester, U.K., and the PhD.
degree from the Queen Mary University of London,
London, U.K. 
His main research interests include signal processing,
machine learning, semantic Web technologies, and the Internet of Things. He is currently a Lecturer with the University of Essex, U.K.
\end{IEEEbiography}

\begin{IEEEbiography}[{\includegraphics[width=1in,height=1.5in,clip,keepaspectratio]{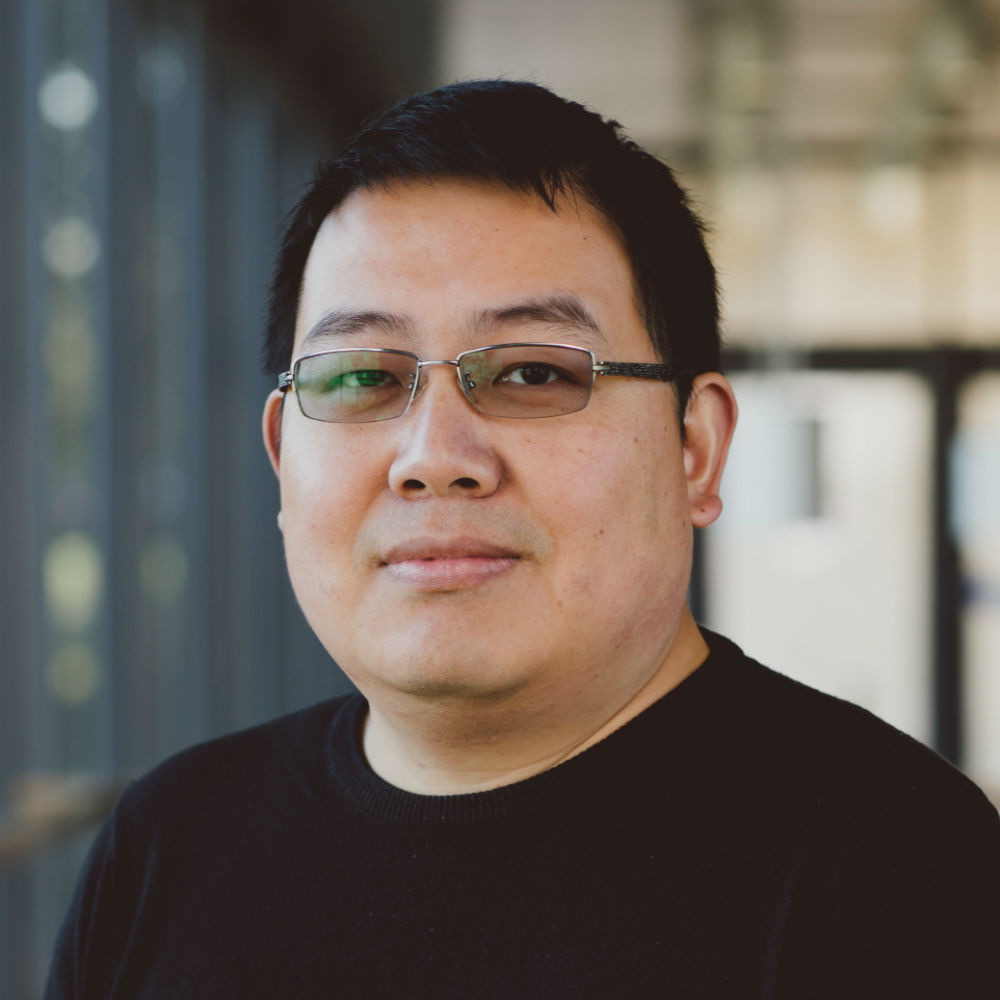}}]{Xiaojun Zhai}  received the B.Sc. degree
from the North China University of Technology,
China, in 2006, and the M.Sc. degree in embedded
intelligent systems and the PhD degree from the
University of Hertfordshire, U.K., in 2009 and
2013, respectively. He is currently a Lecturer with
the School of Computer Science and Electronic
Engineering, University of Essex. His research
interests mainly include the design and implementation
of the digital image and signal processing
algorithms, custom computing using FPGAs, embedded systems, and hardware/software co-design. 
\end{IEEEbiography}

\begin{IEEEbiography}[{\includegraphics[width=1in,height=1.5in,clip,keepaspectratio]{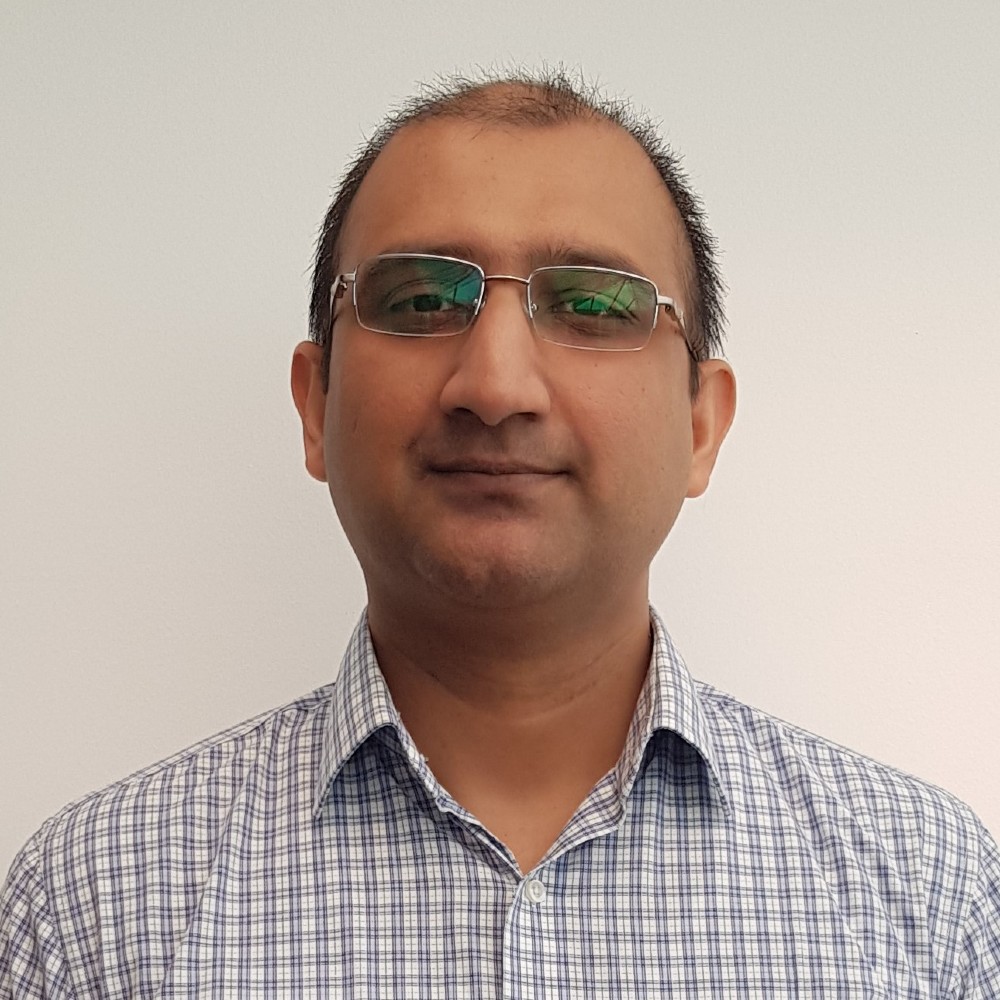}}]{Shoaib Ehsan}  received the B.Sc. degree in
electrical engineering from the University of Engineering
and Technology, Taxila, Pakistan, in
2003, and the PhD degree in computing and electronic
systems (with specialization in computer
vision)from the University of Essex, Colchester,
U.K., in 2012. His current research
interests are in intrusion detection for embedded systems, local feature
detection and description techniques, and image feature matching and performance
analysis of vision systems. 
\end{IEEEbiography}

\begin{IEEEbiography}[{\includegraphics[width=1in,height=1.5in,clip,keepaspectratio]{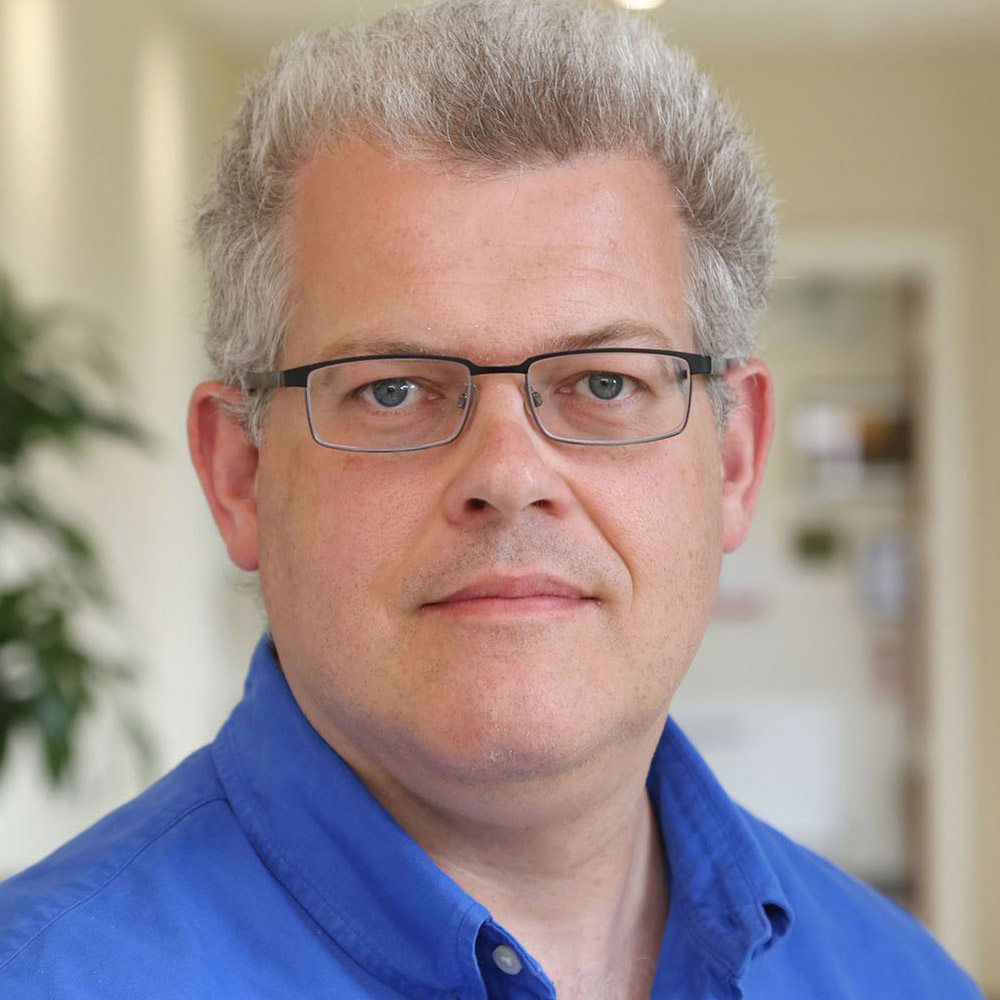}}]{Klaus D. McDonald-Maier}  (S’91–SM’06)is
currently the Head of the Embedded and Intelligent
Systems Laboratory, University of Essex,
Colchester, U.K. He is also the Chief Scientist
with UltraSoC Technologies Ltd., the CEO of
Metrarc Ltd., and a Visiting Professor with the
University of Kent. His current research interests
include embedded systems and system-on-chip
design, security, development support and
technology, parallel and energy-efficient architectures,
computer vision, data analytics, and the application of soft computing
and image processing techniques for real-world problems. 
\end{IEEEbiography}

\end{document}